\documentclass[lettersize,journal]{IEEEtran}
\usepackage{amsmath,amsfonts}
\usepackage{algorithmic}
\usepackage{array}
\usepackage[caption=false,font=footnotesize,labelfont=rm,textfont=rm]{subfig}
\usepackage{textcomp}
\usepackage{stfloats}
\usepackage{mathrsfs}
\usepackage{url}
\usepackage{verbatim}
\usepackage{graphicx}
\usepackage{color}
\usepackage{amssymb}
\def\BibTeX{{\rm B\kern-.05em{\sc i\kern-.025em b}\kern-.08em
    T\kern-.1667em\lower.7ex\hbox{E}\kern-.125emX}}
\usepackage{balance}
\usepackage{amsthm,amsmath,amssymb}
\usepackage{multirow}
\usepackage{booktabs}
\usepackage{makecell}
\usepackage{subfloat} 
\usepackage{cite}
\usepackage{hyperref}
\usepackage{supertabular}
\usepackage{threeparttable}

\hypersetup{hypertex=true,
	colorlinks=true,
	linkcolor=blue,
	anchorcolor=blue,
	citecolor=blue}

\begin{document}
\title{4DRVO-Net: Deep 4D Radar--Visual Odometry Using Multi-Modal and Multi-Scale Adaptive Fusion}
\author{Guirong Zhuo; Shouyi Lu; Huanyu Zhou; Lianqing Zheng; Mingyu Zhou; Lu Xiong$^{\ast}$
	\thanks{
		\emph{(Corresponding author: Lu Xiong).}}
	\thanks{Guirong Zhuo, Shouyi Lu, Huanyu Zhou, Lianqing Zheng, and Lu Xiong are with the School of Automotive Studies, Tongji University, Shanghai 201804, China (e-mails: 2210803@tongji.edu.cn).}
\thanks{Mingyu Zhou is with the Shanghai Geometrical Perception and Learning Co., Ltd., Shanghai 201804, China.}} 
\markboth{}%
{Shell \MakeLowercase{\textit{et al.}}: Sample article using IEEEtran.cls for IEEE Journals}

\maketitle

\begin{abstract}
Four-dimensional (4D) radar--visual odometry (4DRVO) integrates complementary information from 4D radar and cameras, making it an attractive solution for achieving accurate and robust pose estimation. However, 4DRVO may exhibit significant tracking errors owing to three main factors: 1) sparsity of 4D radar point clouds; 2) inaccurate data association and insufficient feature interaction between the 4D radar and camera; and 3) disturbances caused by dynamic objects in the environment, affecting odometry estimation. In this paper, we present 4DRVO-Net, which is a method for 4D radar--visual odometry. This method leverages the feature pyramid, pose warping, and cost volume (PWC) network architecture to progressively estimate and refine poses. Specifically, we propose a multi-scale feature extraction network called Radar-PointNet++ that fully considers rich 4D radar point information, enabling fine-grained learning for sparse 4D radar point clouds. To effectively integrate the two modalities, we design an adaptive 4D radar--camera fusion module (A-RCFM) that automatically selects image features based on 4D radar point features, facilitating multi-scale cross-modal feature interaction and adaptive multi-modal feature fusion. In addition, we introduce a velocity-guided point-confidence estimation module to measure local motion patterns, reduce the influence of dynamic objects and outliers, and provide continuous updates during pose refinement. We demonstrate the excellent performance of our method and the effectiveness of each module design on both the VoD and in-house datasets. Our method outperforms all learning-based and geometry-based methods for most sequences in the VoD dataset. Furthermore, it has exhibited promising performance that closely approaches that of the 64-line LiDAR odometry results of A-LOAM without mapping optimization.
\end{abstract}


\section{Introduction} 
Odometry estimation is a critical technology for autonomous driving, providing high-precision localization for intelligent vehicles in the absence of GPS. This task involves using consecutive images or point clouds to determine the relative pose transformation between two frames. Traditional methods tackle this challenge through a pipeline comprising feature extraction, matching, motion estimation, and optimization\cite{orbslam3,dmvio,lvisam,fastlio2,sdvloam}. Although these approaches have demonstrated excellent performance, they exhibit limitations in handling challenging scenarios. Recently, deep learning techniques have gained widespread application in odometry tasks, showcasing impressive performance in dealing with sparse features and dynamic environments. Within the domain of deep learning-based approaches, numerous studies \cite{ccvo,deepvo,lodonet,lonet} utilize single-modal sensors and construct deep learning models for the purpose of estimating odometry. In addition, a few studies \cite{li2021self,hvlo} employ multi-sensor fusion techniques to accomplish the task in a data-driven manner, leveraging the complementary features of each sensor. The primary objective of this paper is to estimate odometry using a multi-sensor fusion method.
\begin{figure}[!t]
	\centering
	\includegraphics[height=9.5 cm]{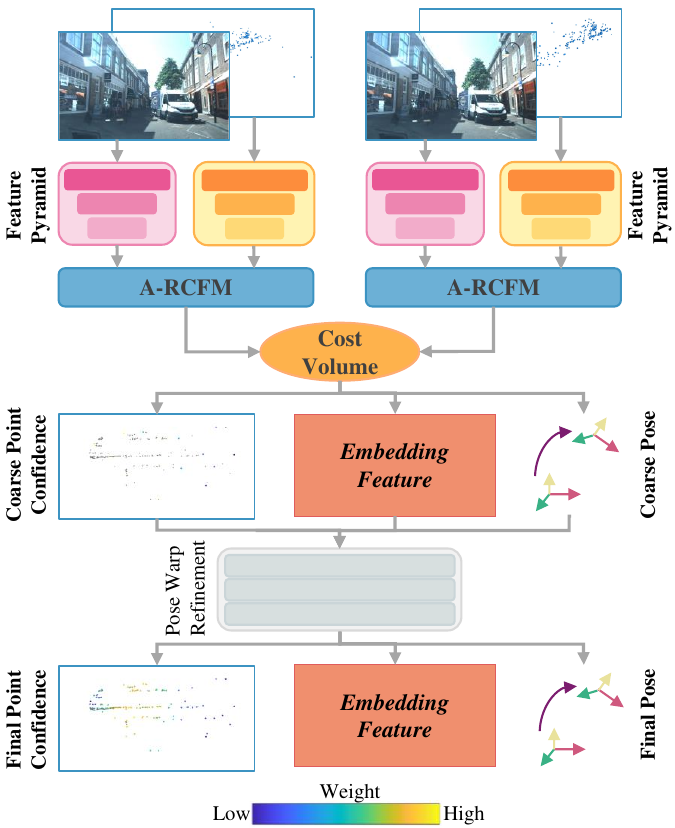}
	\caption{Our proposed 4DRVO-Net combines 4D radar and camera information through the adaptive 4D radar-camera fusion module. We achieve layer-by-layer refinement of pose, point confidence, and embedding features through iterative pose warping.}
	\label{fig1}
\end{figure}

The monocular camera is a widely used sensor in intelligent vehicles due to its compact size, cost-effectiveness, and capability to capture rich semantic information. However, deep visual odometry necessitates relatively stable lighting conditions, predominantly static scenes, and sufficient texture to extract features for accurate camera motion estimation. Moreover, when relying solely on monocular images, obtaining the absolute scale directly is not feasible.Recently, the 4D millimeter-wave radar has garnered increasing attention from both academic and industrial communities\cite{rcfusion,raflow,cmflow,iRIOM}, owing to its distinctive advantages over cameras. Firstly, the 4D radar can directly perceive dynamic objects within the scene, thereby mitigating the adverse effects of dynamic elements on odometry. Secondly, it exhibits robust performance in adverse environmental conditions, including rain, snow, fog, and challenging lighting situations. Lastly, the 4D radar excels in capturing precise spatial information of the surroundings, enabling odometry estimation with absolute scale. However, the main challenges in utilizing 4D radar point clouds for odometry tasks lie in their sparsity and noise.

4D radar and monocular cameras exhibit strong complementarity. Effectively fusing information from these sensors endows sparse radar point clouds with richer feature expression and ensures robustness of the odometry system under various weather conditions. Moreover, the low cost and small size of 4D radar and cameras make them easily applicable in intelligent vehicles with limited budgets and assembly space. Therefore, research on 4D radar visual odometry is of significant value for academic investigation and industrial applications.

Fusing information from monocular cameras and 4D radar for odometry estimation poses three key challenges. Firstly, 4D radar point clouds are sparser, noisier, and have lower resolution compared to the point clouds generated by LiDAR. These characteristics significantly hinder the robust extraction of point cloud features. Secondly, achieving accurate association and full fusion of point features from 4D radar with image pixel features presents a challenge. Lastly, points or pixels associated with dynamic objects in 4D radar point clouds or images are unsuitable for estimating pose. This occurs because these points or pixels violate the assumption of a static environment.

To address the above challenges, in this study, we present a novel deep learning-based 4D radar-visual odometry method named 4DRVO-Net, aiming to tackle the aforementioned challenges. The 4DRVO-Net architecture is constructed based on Feature Pyramid, Pose Warping, and Cost Volume (PWC) structure. Firstly, for efficient utilization of the information-rich 4D radar point clouds, we propose a multi-scale feature extraction network called Radar-PointNet++, leveraging the PointNet++ architecture. The network adeptly extracts point cloud features at various scales by separately encoding and fusing the diverse information from 4D radar data. Secondly, to seamlessly integrate the 4D radar and visual information, we construct image pyramids and point cloud pyramids to derive multi-scale features from both modalities. Subsequently, we design a 4D radar-camera adaptive fusion module using deformable attention, facilitating multi-scale cross-modal feature interaction and adaptive fusion. Finally, we develop a velocity-guided point confidence estimation module to mitigate noise and dynamic object interference, utilizing the velocity information of 4D radar points.

The contributions of this study are as follows.

\begin{itemize}
	\item{Utilizing the feature pyramid, pose warping, and cost volume (PWC) structure, we develop an efficient fully end-to-end framework for 4D radar--visual odometry, which we call 4DRVO-Net. 4DRVO-Net employs a coarse-to-fine hierarchical optimization approach for iterative refinement and accurate estimation of inter-frame poses.}
	\item{In the proposed framework, we introduce a multi-scale feature extraction network called Radar-PointNet++ specifically designed for 4D radar point clouds. This network enables fine-grained learning of sparse point clouds. Moreover, we design an adaptive 4D radar--camera fusion module (A-RCFM) based on deformable attention, facilitating multi-scale cross-modal feature interaction and adaptive multi-modal feature fusion.}
	\item{In the proposed framework, we propose a velocity-guided point confidence estimation method based on the velocity information of 4D radar points. This method aims to model point uncertainty and mitigate the adverse effects of dynamic objects and noise on pose estimation in the environment.}
	\item{Finally, our methods are extensively evaluated using the View-of-Delft (VoD) dataset \cite{vod} as well as the in-house dataset. We conduct evaluation experiments and ablation studies to demonstrate the superiority of the proposed method and the effectiveness of each design.}
\end{itemize}

The remainder of this paper is organized as follows. Section \ref{section2} presents related work. The details of the proposed methodology are presented in Section \ref{section3}. Datasets, baselines, evaluation metrics, and training details are presented in Section \ref{section4}. Experimental results that compare the proposed method with other methods and ablation studies are presented in Section \ref{section5}. Finally, conclusion is drawn in Section \ref{section6}.

\section{Related Work}
\label{section2}
\subsection{Deep Visual Odometry}
Deep learning-based visual odometry has demonstrated superior performance over traditional methods in terms of algorithm robustness, information comprehension, storage, and cognitive patterns \cite{li2022overview}. Consequently, it has garnered significant attention from researchers. The pioneering work by \cite{konda2015learning} introduced the application of deep neural networks for odometry estimation. It initially explores the application of deep learning in this particular field by predicting both the speed and direction of individual images. PoseNet \cite{posenet} utilizes a Convolutional Neural Network (CNN)-based architecture to extract features from the input image and estimate the camera's pose. On the other hand, GoogleLeNet \cite{szegedy2015going} replaces PoseNet's linear regression and Softmax layers used for classification with a fully connected layer that outputs a seven-dimensional pose vector. Furthermore, GoogleLeNet incorporates effective pose supervision by employing the reprojection error as a loss function. DeepVO \cite{deepvo} adopts deep recurrent neural networks to capture the temporal dynamics and interdependencies among sequences, enabling ego-motion estimation. ESP-VO \cite{espvo} extends the work of DeepVO by integrating both pose estimation and uncertainty assessment within a single cohesive framework, allowing for enhanced ego-motion inference while considering associated uncertainties. TartanVO \cite{tartanvo} improves the generalization of the VO model by introducing cosine similarity loss and normalized distance loss, while also integrating camera internal parameters. Additionally, Xue et al. \cite{xue2019beyond} proposed an adaptive memory module that effectively preserves and adapts information from local to global scales in a neural memory simulation, empowering the VO model to effectively handle long-term dependencies. To mitigate the adverse effects of dynamic objects in the environment on the visual odometry, Liu et al. \cite{liu2021unsupervised} computed confidence values for the input image pixels by evaluating the relative similarity of geometrically corresponding regions in the correlated images. DPVO \cite{teed2022deep} employed sparse patch-based matching, instead of dense streams, to achieve optimal accuracy and efficiency. Moreover, Li et al. \cite{cross-modal} developed a visual odometry method based on cross-modal knowledge distillation, where trained visual-LiDAR odometry were used as teachers to guide the training of the VO network. CCVO \cite{ccvo} utilizes two cascaded CNNs for an end-to-end pose estimation. The first CNN is responsible for detecting trackable feature points and performing semantic segmentation to discard feature points belonging to dynamic objects. The second CNN takes static feature points from consecutive images as input and predicts the pose between the frames.
\subsection{Deep Point Cloud Odometry}
In contrast to conventional millimeter-wave radar, 4D radar produces a point cloud that resembles LiDAR rather than a single point target. Deep odometry methods based on LiDAR point clouds have been extensively investigated. Nicolai et al. \cite{nicolai2016deep} pioneered the application of deep learning techniques to point cloud odometry by projecting 3D point clouds onto a 2D plane and subsequently employing image-based deep learning methods for implementing deep point cloud odometry. DeepPCO \cite{deeppco} adopts a panoramic depth view to depict the point cloud and employs a two-branch architecture for distinct translation and rotation estimation. LO-Net \cite{lonet} enhances odometry pose estimation accuracy through the learning of normals and dynamic area masks. LodoNet \cite{lodonet} takes the 3D point clouds and transforms them into the image space, employing an image-based approach for feature extraction and matching to estimate odometry. DMLO \cite{dmlo} addresses the 3D point cloud odometry task by breaking down the 6-DoF pose estimation into two primary components: a matching network based on learning and a pose estimation relying on singular value decomposition. Xu et al. \cite{xu2022robust} proposed a two-stage odometry estimation network for obtaining ego-motion. The first stage involves estimating a set of sub-regional transformations, which are then averaged using a motion voting mechanism. Additionally, they designed a 3D point covariance estimation module to mitigate the interference of moving objects with the odometry by estimating the 3D point covariance. PWCLO-Net \cite{wang2021pwclo} constructs a PWC structure for the 3D point cloud odometry task. The method refines the estimated poses hierarchically, progressively from coarse to fine levels. Moreover, it introduces a trainable embedding mask to evaluate the local motion patterns of all points. EfficientLO-Net \cite{wang2022efficient} introduces a novel approach for representing 3D point clouds, which incorporates projection-awareness. By organizing the raw 3D point clouds into an ordered data form, the efficiency of point cloud odometry is significantly enhanced. NeRF-LOAM \cite{deng2023nerf} incorporates neural radiation fields into the 3D point cloud odometry system, exhibiting robust generalization capabilities across various environments. STCLoc \cite{yu2022stcloc} unites absolute positional regression with a pioneering classification task, which categorizes point clouds based on position and orientation, aiming to standardize the regression of positional poses. Deep odometry methods utilizing 4D radar point clouds have not undergone comprehensive investigation, and only a limited number of studies have addressed the task of estimating 4D radar point cloud odometry as an intermediate step \cite{raflow,cmflow}. RaFlow \cite{raflow} is centered around 4D radar-based scene flow estimation, with odometry as an intermediate task. On the other hand, CMFlow \cite{cmflow} is a 4D radar-based method that incorporates cross-modal learning to estimate scene flow, motion segmentation, and odometry. The odometry task is effectively supervised using GPS/INS data.
\subsection{Visual-LiDAR Odometry}
In prior research, the odometry system that fuses visual and point cloud data, known as visual-LiDAR odometry, makes use of both sensor modalities to achieve functional complementarity. DEMO \cite{zhang2014real} employed LiDAR point clouds to provide depth information to the RGB images. V-LOAM \cite{zhang2015visual} utilized the estimated poses obtained from visual odometry at high frequencies as a motion prior for low frequency LiDAR odometry, leading to the refinement of the estimated motion. \cite{huang2020lidar} leveraged additional information about the environment's structure by incorporating point and line features into the pose estimation process. Moreover, LiDAR point clouds were employed to provide depth information for both points and lines. LIMO \cite{limo} employed the depth information obtained from LiDAR to mitigate the scale uncertainty inherent in monocular visual odometry. SDV-LOAM \cite{sdv} integrates a semi-direct visual odometry with an adaptive sweep-to-map LiDAR odometry to address challenges, including the 3D-2D depth correlation in visual-LiADR odometry. In recent times, several visual-LiDAR odometry methods have emerged, all of which are founded on deep learning principles. Yue et al. \cite{yue2020lidar} introduced a novel deep learning approach to enhance raw 3D point clouds obtained from a low-cost 16-channel LiDAR system using high-resolution camera images. Subsequently, the enhanced point clouds were employed in LiDAR SLAM, which was based on the Normal Distribution Transform (NDT) algorithm. H-LVO \cite{hvlo} is a novel hybrid visual-LiDAR odometry framework that utilizes depth maps, depth flow maps, and deep LiDAR depth complementation networks to facilitate 2D feature matching and 3D correlation. Additionally, a geometry-based approach is employed for pose solving. Self-VLO \cite{li2021self} is a self-supervised framework for visual-LiDAR odometry, which utilizes monocular images and sparse depth maps acquired through 3D LiDAR point projection as its inputs. The framework incorporates a dual-path encoder for extracting features from both visual and depth images. MVL-SLAM \cite{an2022visual} adopts the RCNN network architecture and takes RGB images and multi-channel depth images as inputs. These depth images are generated from 3D LiDAR point clouds, and the framework produces poses with an absolute scale.

\section{4DRVO-Net}
\label{section3}
\subsection{Overview}

\begin{figure*}[tb]
	\centering
	\includegraphics[width=0.95\linewidth]{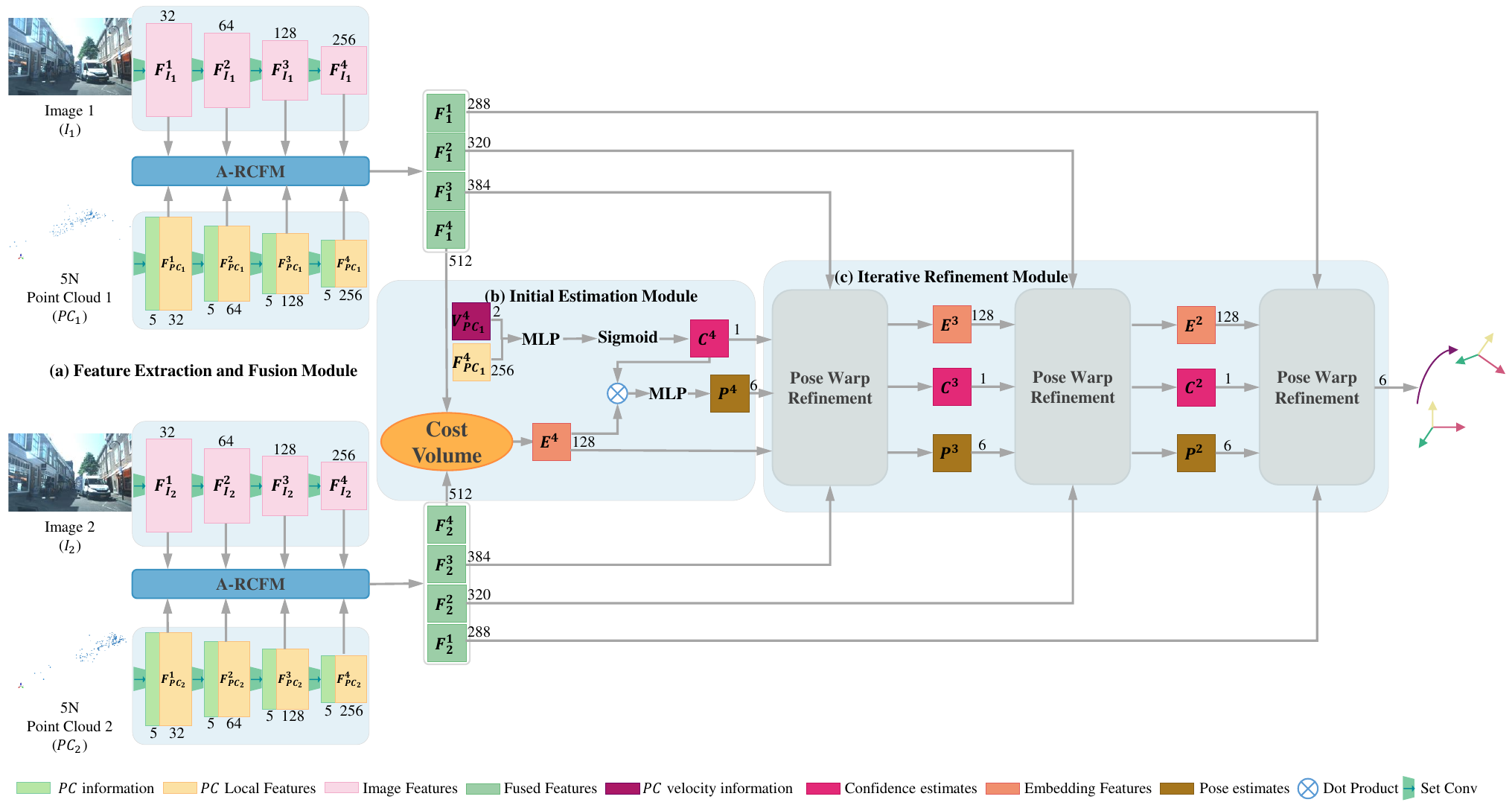}
	\centering
	\caption{The 4DRVO-Net architecture is proposed, comprising three main components: (a) A feature extraction and fusion module, which consists of multi-scale feature extraction blocks and an adaptive 4D radar--camera fusion block. This module is utilized to extract RGB image features and 4D radar point features at four different scales and then fuse them together in an adaptive manner at each scale. (b) An initial estimation module is employed to associate two frames of 4D radar point clouds with fused features and to perform an initial estimation of point confidence and pose. (c) Additionally, an iterative refinement module is incorporated to finely adjust the point confidence, embedding features, and pose, progressively improving their accuracy.}
	\label{fig2}
\end{figure*}

This study introduces 4DRVO-Net, a novel end-to-end learning framework for 4D radar--visual odometry. The overall architecture of the proposed 4DRVO-Net architecture is illustrated in Fig. \ref{fig2}. The network takes two pairs of synchronized inputs: RGB images and 4D radar point clouds, denoted as ($I_1 \in \mathbb{R}^{H \times W \times 3}$, $PC_1=\left\{x_i\in \mathbb{R}^5|i=1,...,N_1\right\}$), and ($I_2 \in \mathbb{R}^{H \times W \times 3}$, $PC_{2}=\left\{y_j\in \mathbb{R}^5|j=1,...,N_2\right\}$) respectively. We begin by encoding these inputs using the image and point-cloud feature pyramids. The feature pyramid network comprises four sets of convolutional layers, each performing downsampling on the input images and point clouds, while extracting local features. Moreover, these convolutional layers share identical weights at the same level. Subsequently, at each scale, we propose an adaptive 4D radar--camera fusion module (A-RCFM) that effectively associates synchronized 4D radar and camera frames to enhance the extremely sparse 4D radar features that lack discriminative details. Detailed information regarding this module is provided in Section \ref{3.2}. Next, we utilize a cost--volume network to associate 4D radar frames with fused features to generate embedding features. In addition, we perform an initial estimation of the point confidence, which influences the embedding of features back into the initial poses. Detailed information for this process is provided in Section \ref{3.3}. In Section \ref{3.4}, we introduce an Iterative Refinement Module to progressively refine the embedding features, point confidence, and pose from coarse to fine. Finally, the network outputs the Euler angles ($eula \in \mathbb{R}^3$) and translation vectors ($t \in \mathbb{R}^3$) between consecutive radar frames.
\subsection{Multi-scale Feature Extraction and Adaptive Feature Fusion}
\label{3.2}
\subsubsection{Multi-scale Feature Extraction}
The multi-scale feature extraction block comprises two symmetric branches, namely, the image branch and point branch, tailored for 2D and 5D data, respectively. Both branches are equipped with Siamese pyramids comprising multiple sets of convolutional layers for multi-scale feature encoding and extraction. For the image branch, we employ an image pyramid network consisting of residual blocks \cite{he2016deep} to extract features from the RGB images. The extracted image features at various scales can be represented as $F_I^1$, $F_I^2$, $F_I^3$, and $F_I^4$ based on the forward direction indicated in Fig. \ref{fig2}.

\begin{figure}[!t]
	\centering
	\includegraphics[width=0.98\linewidth]{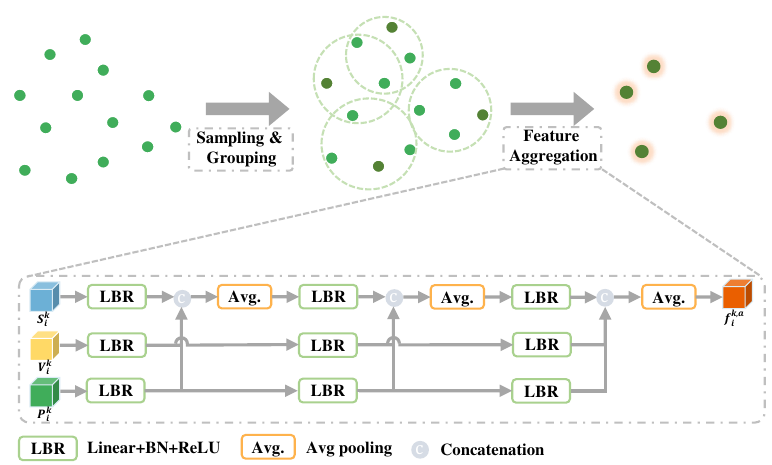}
	\caption{The Radar-PointNet++ structure comprises the following steps: Initially, the input point cloud undergoes sampling and grouping using farthest point sampling (FPS) and the K-nearest neighbors algorithm (KNN). Afterward, the spatial, velocity, and intensity information of radar points in each group is individually encoded. A deep fusion process is then applied to combine the three feature scales, resulting in the computation of the aggregated point features.}
	\label{fig3}
\end{figure}

Four-dimensional radar point clouds are frequently characterized by sparsity, noise, and uneven distribution. Although expanding the receptive field can mitigate the sparsity problem in point clouds, performing feature extraction only at a single scale is inadequate for tackling the challenge of unevenly distributed point clouds. Drawing inspiration from the ‘‘hierarchical point set feature learning'' component of PointNet++ \cite{pointnet++}, we present Radar-PointNet++, a novel 4D radar point cloud pyramid network. This architecture allows for multi-scale feature encoding of 4D radar point clouds, as illustrated in Fig. \ref{fig3}. At each scale, the input comprises $n$ points $\left\{p_i=\left\{x_i,f_i\right\}|i=1,2,...,n\right\}$, where each point is characterized by 5D point information $x_i\in\mathbb{R}^5$ and the feature $f_i\in \mathbb{R}^c$ derived from the previous scale. The output consists of $n^{\prime}$ sampled points $\left\{p_j^{\prime}=\left\{x_j^{\prime},f_j^{\prime}\right\}|j=1,2,...,n^{\prime}\right\}$, where $x_j^{\prime}\in\mathbb{R}^5$ represents the 5D point information, and $f_j^{\prime} \in \mathbb{R}^c$ corresponds to the extracted local features. These sampled points are obtained from the input point cloud using farthest point sampling (FPS).

For each sampled point $p_j^{\prime}$, we utilize the K-nearest neighbors algorithm (KNN) to select $K$ neighboring points $\left\{p_j^{k}=\left\{x_i^{k},f_i^{k}\right\}|k=1,2,...,K\right\}$ from the input point cloud. Here, $x_i^k$ and $f_i^k$ represent the 5D information and point features, respectively, for each neighboring point $p_j^{k}$. Subsequently, the newly devised feature aggregation network performs feature extraction on these $K$ points. The spatial coordinate information $S \in \mathbb{R}^3$, velocity information $V \in \mathbb{R}^1$, and intensity information $P \in \mathbb{R}^1$ of 4D radar points provide distinct environmental characteristics. We consider that mapping all information from 4D radar points to the same feature space could lead to feature confusion. Therefore, for $K$ neighboring points, their spatial, velocity, and intensity information are individually learned through three separate linear layers with distinct weights, followed by batch normalization (BN) and rectified linear unit (ReLU) activation. Deep-feature fusion is then performed at three feature scales. Subsequently, the aggregated point features $f_i^{k,a}$ are computed using an average pooling operation. We connect $f_i^{k,a}$, $f_i^k$, $S_i^k$, $V_i^k$, and $P_i^k$, and after passing through a linear layer, BN layer, and ReLU layer, we obtain the final local features $f_j^{\prime}$. The formula used is as follows:
\begin{equation}
	\label{equation1}
	\begin{aligned}
		f_j^{\prime}=&LBR(f_i^k \oplus Aggregation(S_i^k,V_i^k,P_i^k)\oplus S_i^k \\
		             & \oplus V_i^k \oplus P_i^k) (k=1,...,K),\\
	\end{aligned}
\end{equation}
where $\oplus$ denotes the concatenation of two vectors, $LBR$ refers to the linear, BN, ReLU layers, and $Aggregation$ represents the aggregated feature network block described above.

In this study, we sample $N$ points from the original 4D radar point cloud and extract $N/2$, $N/4$, $N/8$, and $N/16$ points and their corresponding local features in a hierarchical manner from the sampled 4D radar point set. Based on the directional markers indicated in Fig. \ref{fig2}, the coordinate sets of 4D radar points at different scales is denoted as $PC^1$, $PC^2$, $PC^3$, and $PC^4$, and their corresponding local features are denoted as $F_{PC}^1$, $F_{PC}^2$, $F_{PC}^3$, and $F_{PC}^4$, respectively.
\subsubsection{Adaptive Feature Fusion}
\begin{figure}[!t]
	\centering
	\includegraphics[width=0.98\linewidth]{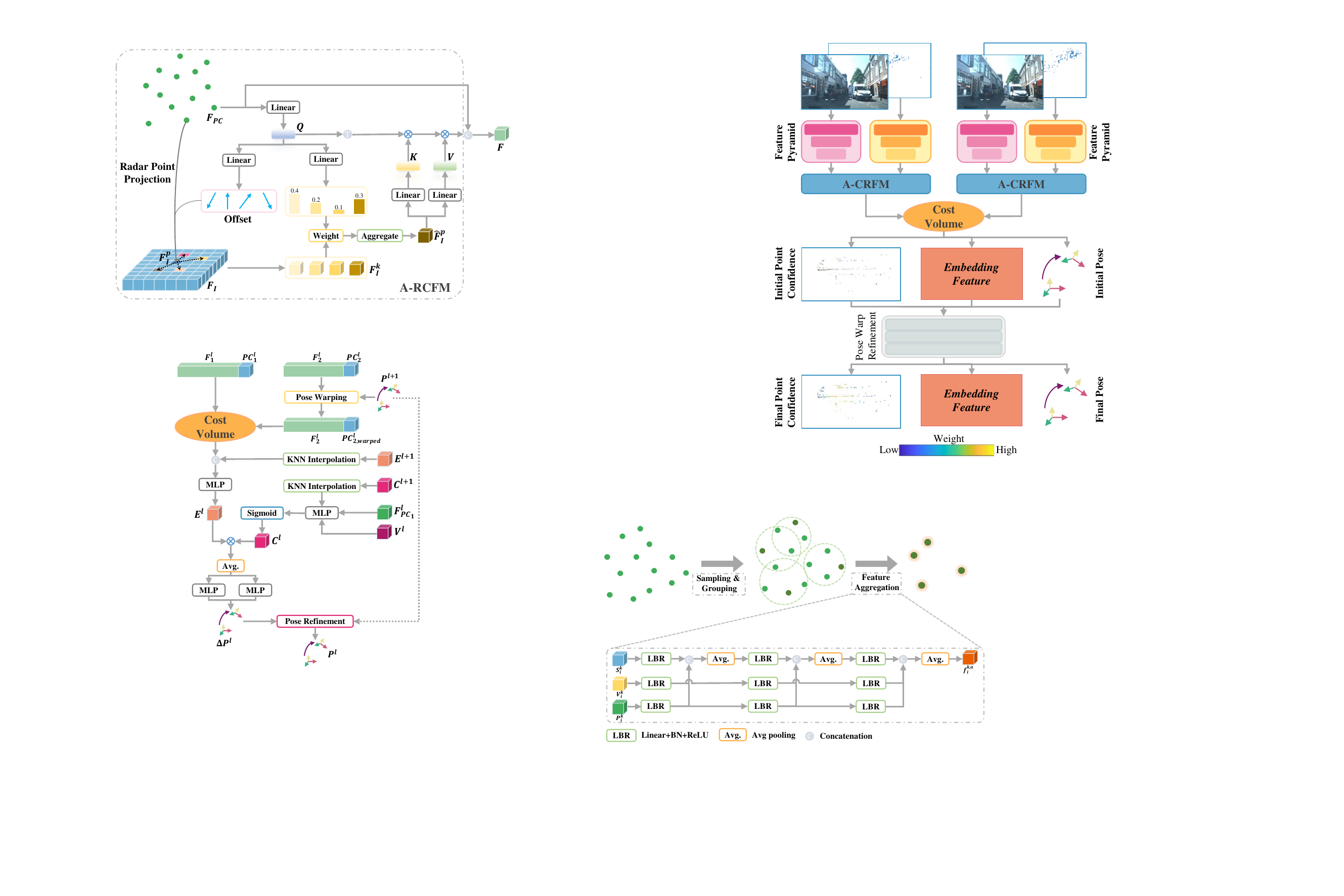}
	\caption{Adaptive 4D radar--camera fusion module. Firstly, we project the 4D radar points onto the image plane. Using learnable dynamic offsets, we seek the semantic features $F_I^k$ in the image feature $F_I$ that are relevant to the current 4D radar point features. The semantic features are then aggregated using weights. Then, we use the cross-attention module to fuse the aggregated semantic features $\widehat{F}_I^p$ and the point features $F_{PC}$, generating the adaptive fusion features $F_{I\leftarrow P}$. Finally, we connect the adaptive fusion feature $F_{I\leftarrow P}$ with the original point feature $F_{PC}$ to obtain the fusion feature $F$.} 
	\label{fig4}
\end{figure}
Precisely associating and fully interacting 4D radar point features with image features is challenging. In this section, we introduce a novel approach for effectively fusing RGB images and 4D radar point cloud features. By leveraging the 4D radar camera external matrix $T_{cr}$ and the camera internal matrix $K$ at different feature scales, we project the 4D radar sample points with coordinates $PC$ and local features $f_j^{\prime}$ from the radar coordinate system onto the feature map of the RGB image. This projection yields the corresponding projected point $p$ with coordinates $(u,v)$ and associated semantic feature $F_I^p$. In cases where the coordinates $(u,v)$ are not integers, bilinear interpolation is employed for retrieval. The projection process is as follows:
\begin{equation}
	\label{equation2}
	\begin{aligned}
		\begin{bmatrix}
			u \\
			v \\
			1 \\
		\end{bmatrix} = \frac{1}{Z}K(T_{cr}PC)_{(1-3)}
	\end{aligned}
\end{equation}
where $PC$ represents the homogeneous coordinates of 4D radar points. After multiplying $T_{cr}$ by $PC$, we select the first three dimensions of $T_{cr}\times PC$ such that they can be multiplied by $K$, which is a $3\times3$ matrix.

It is essential to consider that the projected points $p$ might not align perfectly with the 4D radar points $PC$, owing to potentially small variations in the external parameters resulting from calibration errors or vehicle jitter. Moreover, we argue that sparse 4D radar points should dynamically discover relevant semantic features from a dense image feature map for fusion to achieve a full interaction between the point and image features. To address these challenges, we propose a deformable attention-based spatial cross-attention mechanism in which the 4D radar point feature $F_{PC}$ adaptively identifies regions of interest on the image feature map for interaction. Specifically, we first project $F_{PC}$ onto a point-feature query $Q^P=F_{PC}\cdot W_P$. Subsequently, two linear transformations are applied to $Q^P$, resulting in a dynamic offset $\Delta p_k$ and the corresponding weight $A_k$ for $K$ sampling points. By applying the learned offset $\Delta p_k$ to the image feature map $F_I$ based on projection point $p$, we obtain a set of image features $F_I^k$ that are closely associated with the 4D radar points. The aggregated image features $\widehat{F}_I^p$ are obtained by applying weights $A_k$ to image features $F_I^k$. Next, $\widehat{F}_I^p$ is projected onto the key $K^I=\widehat{F}_I^p\cdot W_K$ and value $V^I=\widehat{F}_I^p\cdot W_V$. The multi-head attention mechanism is then applied to process features $Q^P$, $K^I$, and $V^I$, resulting in the generation of the adaptive fusion feature $F_{I\leftarrow P}$. The parameters $W_K$, $W_P$, and $W_V$ represent learnable linear mapping. The adaptive 4D radar--camera fusion process is formulated as follows:
\begin{equation}
	\label{equation3}
	\begin{aligned}
		&F_I^k=F_I(p+\Delta p_k),\\
		&\widehat{F}_I^p=\sum_{k=1}^{K} A_kF_I^k,\\
		&F_{I\leftarrow P}=MultiheadAttn(Q^P,K^I,V^I)\\
	\end{aligned}
\end{equation}
where $MultiheadAttn()$ represents the multi-head attention block. Finally, we connect the adaptive fusion feature $F_{I\leftarrow P}$ and the original point feature $F_{PC}$ to obtain the fusion feature $F$.
\subsection{Point Cloud Association and Velocity-Guided Point Confidence Estimation}
\label{3.3}
\subsubsection{Point Cloud Association}
\label{3.3.1}
We utilize the cost volume layer proposed by \cite{pointpwc} to establish associations between the two 4D radar point clouds. This layer employs a patch-to-patch approach to effectively mitigate the sparsity issue encountered in 4D radar point clouds. This correlates the two point clouds using fused features, thereby generating a point embedding feature  $E=\left\{e_i|e_i\in \mathbb{R}^c\right\}_{i=1}^n$, which effectively describes the correlation information between the two point clouds.
\subsubsection{Velocity-Guided Point Confidence Estimation}
\label{3.3.2}
Utilizing embedding features $E$ to generate globally consistent poses between two frames presents a novel challenge. In this section, we introduce a velocity-guided point confidence estimation module to derive a  pose transformation from the embedding features.

The 4D radar point cloud contains dynamic objects, noisy points, and obscured points from other frames, all of which can adversely affect odometry estimation. Filtering is essential for retaining only valuable points for odometry tasks. Unlike other sensors, 4D radar points provide unique velocity information that corresponds directly to dynamic objects in the environment. Leveraging this velocity information can effectively diminish the adverse impact of dynamic points in the point cloud on odometry estimates. Therefore, we propose a velocity-guided point confidence estimation module, which first inputs the velocity information $V=\left\{v_i|v_i\in \mathbb{R}^2\right\}_{i=1}^n$ and point features $F_{PC_1}$ of the point cloud $PC_1$ into the multilayer perceptron (MLP) layer and then performs a sigmoid operation to obtain a confidence estimate $C=\left\{c_i|c_i\in \mathbb{R}^1\right\}_{i=1}^n$ for each point. For point velocity information, we employ the absolute value of the relative radial velocity (RRV) for each point in the point cloud. In addition, we consider the absolute difference between the RRV of each point and the median of the absolute RRV values for all points in the point cloud. The inclusion of the point feature $F_{PC_1}$ not only enhances the sensitivity of the point confidence estimation module to dynamic points, but also evaluates the influence of other factors, such as the spatial location of points, on the odometry estimates. The confidence estimation process is as follows:
\begin{equation}
	\label{equation4}
	\begin{aligned}
		&C=Sigmoid(MLP(F_{PC_1} \oplus V)).\\
	\end{aligned}
\end{equation}

In this process, each point in the point cloud is assigned a weight between 0 and 1, representing its confidence level. A lower weight implies a smaller contribution to the odometry estimate, whereas a higher weight indicates a greater contribution. Subsequently, the estimated point confidence $C$ is utilized to apply weighting to embedding feature $E$. Two separate MLP layers are then employed to regress the rotation estimate $eula$ and translation estimate $t$ for two consecutive frames.
\begin{equation}
	\label{equation5}
	\begin{aligned}
		&eula=MLP(AVG(C \odot E))\\
		&t=MLP(AVG(C \odot E)),\\
	\end{aligned}
\end{equation}
where $AVG()$ denotes an average pooling operation.

The point confidence $C$ undergoes continuous refinement in the subsequent pose warp refinement module, as depicted in Fig. \ref{fig2}. The confidence estimate is propagated to the denser layers of the point cloud and optimized in a coarse-to-fine manner, leveraging the estimates from the sparse layers. This process leads to a more accurate calculation of the final confidence estimate and pose transformation.

\subsection{Pose Warp Refinement Module}
\label{3.4}
\begin{figure}[!t]
	\centering
	\includegraphics[width=0.70\linewidth]{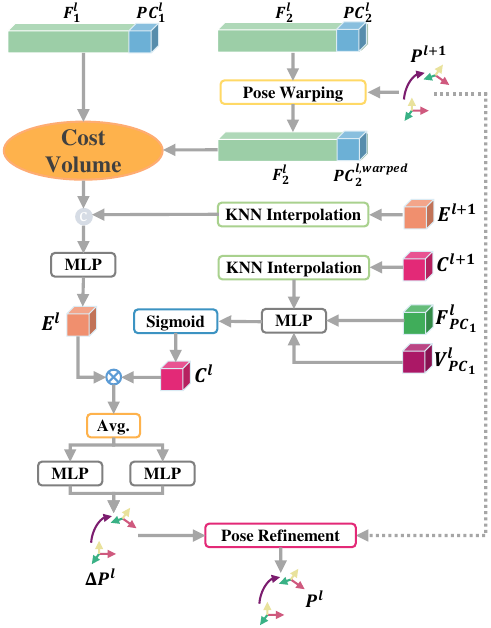}
	\caption{The details of the proposed pose warp refinement module at the $l$-th level. The module takes the embedding features $E^{l+1}$, confidence estimate $C^{l+1}$, and pose estimate $P^{l+1}$ from the $l+1$-th level as inputs, and outputs the corresponding updated embedding features $E^{l}$, confidence estimate $C^{l}$, and pose estimate $P^{l}$ at the $l$-th level after the pose warping refinement.} 
	\label{fig5}
\end{figure}
To achieve the iterative refinement of the point-cloud poses and obtain more accurate pose estimates, we introduce a pose warping refinement module, as depicted in Fig. \ref{fig5}. This module comprises several key components, including a KNN interpolation layer, pose warping, embedding features, point confidence, and pose refinement.
\subsubsection{KNN Interpolation Layer}
To achieve coarse-to-fine refinement of the embedding features and point confidence, we employed the KNN interpolation layer to propagate the aforementioned estimates from sparse to denser layers, serving as the initial values for the refinement process. The input of the KNN interpolation layer comprises the embedding features $E^{l+1}$ and confidence estimate $C^{l+1}$ obtained from the $l+1$-th level. The output consists of the KNN-interpolated embedding features $E^{l,knn}=\left\{e_i^{l,knn}|e_i^{l,knn}\in \mathbb{R}^C\right\}_{i=1}^{n_1^l}$ and the KNN-interpolated confidence estimate $C^{l,knn}=\left\{c_i^{l,knn}|c_i^{l,knn}\in \mathbb{R}^1\right\}_{i=1}^{n_1^l}$. Specifically, each dense point at the $l$-th level selects its K-nearest neighbors from the sparse points at the $l+1$-th level. The distances between the dense points and selected sparse points are then used as weights to aggregate the embedding features and confidence estimates of the chosen sparse points through weighted summation, resulting in KNN-interpolated embedding features and KNN-interpolated confidence estimates.
\subsubsection{Pose Warping Layer}
The pose warping layer is essential for achieving iterative pose refinement. It applies the Euler angles $eula^{l+1}$ and translation vector $t^{l+1}$ from the $l+1$-th level to $PC_2^l=\left\{y_i^l \in \mathbb{R}^3 |i=1,...,N_2^l\right\}$, generating $PC_2^{l,warped}=\left\{y_i^{l,warped} \in \mathbb{R}^3 |i=1,...,N_2^l\right\}$. The equation for pose warping is as follows:
\begin{equation}
	\label{equation6}
	\begin{aligned}
		&y_i^{l,warped}=R^{l+1}y_i^l+t^{l+1},\\
	\end{aligned}
\end{equation}
where $R^{l+1}$ denotes the rotation matrix corresponding to $eula^{l+1}$. If $eula^{l+1}$ and $t^{l+1}$ are sufficiently accurate, $PC^{l}_1$ and $PC_2^{l,warped}$ are infinitely close. However, owing to estimation errors, $PC_2^{l,warped}$ is closer to $PC^{l}_1$ than $PC^{l}_2$. Consequently, the pose refinement process involves estimating the residual motion of $PC_2^{l,warped}$ and $PC^{l}_1$ at the $l$-th level and subsequently correcting $eula^{l+1}$ and $t^{l+1}$ to achieve more accurate pose estimates.
\subsubsection{Embedding Feature and Point Confidence Refinement}
The point clouds $PC_2^{l,warped}$ and $PC^{l}_1$, along with their associated fused features $F_2^l$ and $F_1^l$, are fed as inputs into the cost volume layer, as described in Section \ref{3.3.1}. This process results in the computation of the coarse embedding features $E^{l,coarse}=\left\{e_i^{l,coarse}|e_i^{l,coarse}\in \mathbb{R}^C\right\}_{i=1}^{n_1^l}$ for the $l$-th level. Subsequently, the KNN-interpolated embedding features $e_i^{l,knn}$ and coarse embedding features $e_i^{l,coarse}$ are concatenated and fed into the MLP to obtain the refined embedding features $E^{l}=\left\{e_i^{l}|e_i^{l}\in \mathbb{R}^C\right\}_{i=1}^{n_1^l}$ in the $l$-th layer:
\begin{equation}
	\label{equation7}
	\begin{aligned}
		&e_i^{l}=MLP(e_i^{l,knn} \oplus e_i^{l,coarse}).\\
	\end{aligned}
\end{equation}

To update the point confidence estimates, similar to refining the embedding features, we concatenate the local features $F_{PC_1}^l=\left\{f_{pc,i}^l|f_{pc,i}^l\in \mathbb{R}^C\right\}_{i=1}^{n_1^l}$, point velocity information $V^l=\left\{v_{i}^l|v_{i}^l\in \mathbb{R}^2\right\}_{i=1}^{n_1^l}$, and confidence estimates $C^{l,knn}$ obtained from KNN interpolation. Subsequently, MLP and sigmoid operations are performed on the concatenated features to obtain the refined point-confidence estimates $C^{l}=\left\{c_i^{l}|c_i^{l}\in \mathbb{R}^1\right\}_{i=1}^{n_1^l}$ in the $l$-th layer.
\begin{equation}
	\label{equation8}
	\begin{aligned}
		&c_i^{l}=Sigmoid(MLP(f_{pc,i}^l \oplus v_i^l \oplus c_i^{l,knn})).\\
	\end{aligned}
\end{equation}
\subsubsection{Pose Refinement}
Following the approach described in Section \ref{3.3.2}, we calculate the weighted embedding feature $E^l$ using the estimated point-confidence $C^l$. The residual transformations $\Delta eula^l$ and $\Delta t^l$ between $PC_2^{l,warped}$ and $PC^{l}_1$ can be obtained using two separate MLP layers. Applying these transformations to $eula^{l+1}$ and $t^{l+1}$ yields refined poses $eula^{l}$ and $t^{l}$ in the $l$-th layer:
\begin{equation}
	\label{equation9}
	\begin{aligned}
		&R^{l}=\Delta R^l R^{l+1}\\
		&t^{l}=\Delta R^l t^{l+1}+\Delta t^l.\\
	\end{aligned}
\end{equation}
\subsection{Training Loss}
The network estimates the pose transformations between consecutive frames in the four layers. For pose estimation in each layer, we employ the rotational loss $L_e$ and translational loss $L_t$ to learn the rotational and translational components of pose transformations, respectively.
\begin{equation}
	\label{equation10}
	\begin{aligned}
		&L_{e}=\left\|eula^{gt}-eula^{l} \right\|_2,\\
		&L_{t}=\left\|t^{gt}-t^{l} \right\|_2,\\
	\end{aligned}
\end{equation}
where $\left\|\cdot\right\|$ denotes the $L2$ norm; $eula^{gt}$ and $t^{gt}$ are the ground-truth Euler angle and translation vector, respectively, generated from the ground-truth pose transformation matrix; and $eula^{l}$ and $t^{l}$ are the estimated Euler angles and translational vectors of the network at different layers, respectively.

Owing to the difference in scale and units between the Euler angle and translational vector, referring to a previous study on deep odometry \cite{lonet}, we introduce two learnable parameters, $s_e$ and $s_t$. For each layer, the training loss function is
\begin{equation}
	\label{equation11}
	L=L_e exp(-s_e)+s_e+L_t exp(-s_t)+s_t.
\end{equation}

Thus, under multilayer supervision, the total loss function of the network is
\begin{equation}
	\label{equation12}
	L=\sum^{L}_{l=1} \lambda^l L^l,
\end{equation}
where $L$ is the number of layers, and $\lambda^l$ denotes the weights in the $l$-th layer.

\section{Implementation}
\label{section4}
\subsection{Dataset}
\subsubsection{The View-of-Delft (VoD) automotive dataset}
The View-of-Delft (VoD) automotive dataset \cite{vod} was publicly released in 2022. This dataset encompasses diverse scenes in Delft, including campuses, suburbs, and the old town. It incorporates data from multiple sensors such as cameras, 4D radar, and LiDAR, and also provides the external parameter transformation between different sensors and odometry information. The dataset comprises a total of 8682 frames, organized into 24 sequences based on frame continuity. For training, validation, and testing purposes, we adopt the predefined sets from \cite{cmflow} to evaluate the performance of the network.
\subsubsection{The In-house dataset}
\begin{figure}[tb]
	\centering
	\subfloat[Data Collection Trajectory of Seq.05]{
		\includegraphics[width=0.45\linewidth]{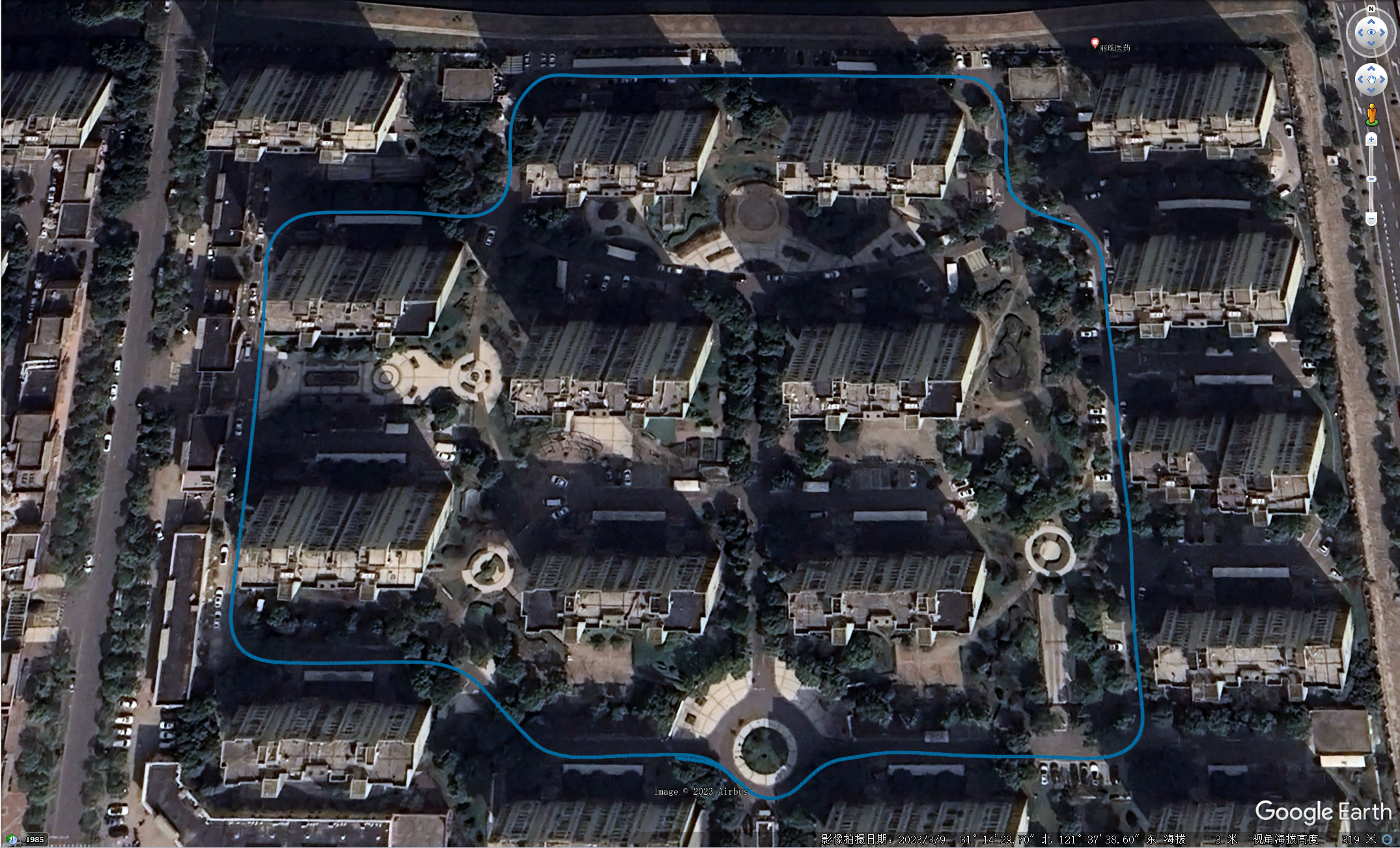}
	}
	\quad
	\subfloat[Data Collection Trajectory of Seq.06]{
		\includegraphics[width=0.45\linewidth]{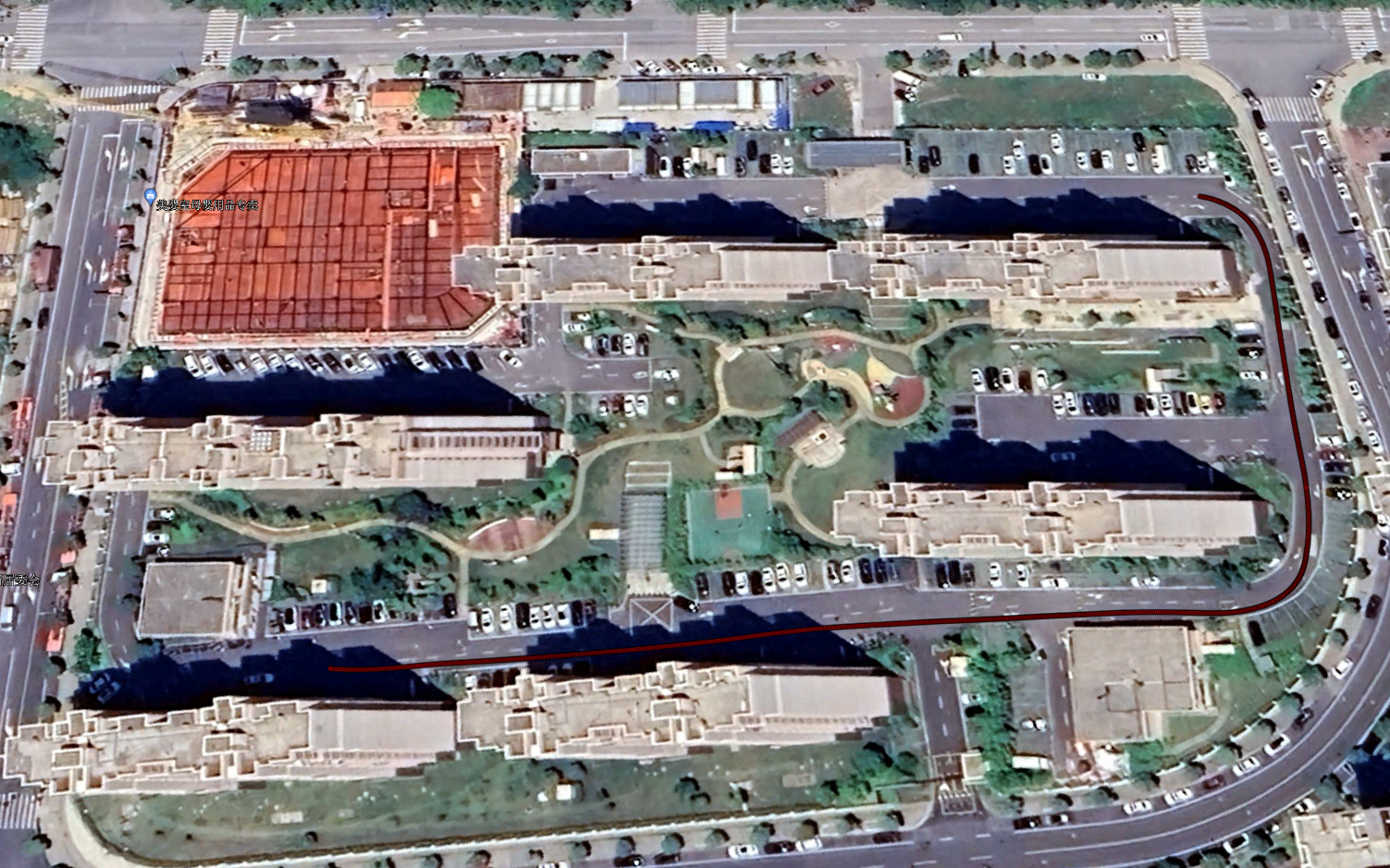}
	}
	\caption{In the In-house dataset, the data collection route of the testing sequence. 
		\label{rount}}
\end{figure}

We utilized a Volkswagen ID4 as our data collection platform, equipped with a 4D radar, a wide-angle camera, and a high-precision multi-source information fusion positioning device. The 4D radar and camera were precisely time-synchronized through hardware triggering, generating data at a frequency of 10Hz. On the other hand, the high-precision multi-source information fusion positioning device operated at a frame rate of 100Hz. Our data collection effort resulted in 7 sequences, encompassing a total of 20,500 frames. To partition the data for training, validation, and testing, we adopted a ratio of 6:2:2. Consequently, we designated sequences 05 and 06, containing 4,100 frames, as the testing set, and sequence 04, comprising 4,300 frames, as the validation set.The data collection route for the testing set is shown in Fig. \ref{rount} The remaining data was allocated for the training set.

\begin{table*}[htbp]
	\renewcommand{\arraystretch}{1.2}
	\caption{The 4D radar-visual odometry experiment results on the VoD dataset. The best result is bolded in red, the second best is bolded in blue.}
	\label{table1}
	\centering
	\resizebox{\textwidth}{!}{
		\begin{tabular}{cc|cc|cc|cc|cc|cc|cc|cc|cc}
			\toprule
			\multicolumn{2}{c|}{\multirow{2}{*}{Method}} & \multicolumn{2}{c|}{03} & \multicolumn{2}{c|}{04} & \multicolumn{2}{c|}{09} & \multicolumn{2}{c|}{17} & \multicolumn{2}{c|}{19} & \multicolumn{2}{c|}{22} & \multicolumn{2}{c|}{24} & \multicolumn{2}{c}{Mean} \\ \cmidrule(l){3-18} 
			& & $t_{rel}$          & $r_{rel}$         & $t_{rel}$          & $r_{rel}$         & $t_{rel}$          & $r_{rel}$         & $t_{rel}$          & $r_{rel}$         & $t_{rel}$          & $r_{rel}$         & $t_{rel}$           & $r_{rel}$  & $t_{rel}$          & $r_{rel}$   & $t_{rel}$          & $r_{rel}$        \\ \cmidrule(r){1-18} 
			\multirow{4}{*}{Classical-based methods} &
			ICP-po2po               & 0.39      & 1.00    & 0.21     & 1.14    & 0.15     & 0.72    & 0.16     & 0.53    & 1.40     & 4.70    & 0.44      & 0.76   & 0.24  & 0.77  & 0.43  & 1.37     \\
			& ICP-po2pl               & 0.42     & 2.19    & 0.37    & 1.83    & 0.50     & 1.32    & 0.23     & 0.68     & 3.04     & 5.62    & 0.42      & 1.20 & 0.35 & 0.67 & 0.76 & 1.93     \\
			& GICP                    & 0.46     & 0.68     & 0.30     & 0.39     & 0.51     & 0.32    & 0.40     & 0.10    & 0.51     & 1.23    & 0.34      & 0.57 & 0.15 & 0.30 & 0.38 & 0.51     \\
			& NDT                     & 0.55     & 1.60    & 0.47     & 0.91     & 0.46     & 0.56    & 0.44     & 0.40    & 1.33     & 2.58    & 0.47      & 1.10 & 0.36 & 1.84 & 0.58 & 1.28 
			\\ \cmidrule(r){1-18}
			\multirow{2}{*}{LiDAR-based methods} &
			A-LOAM w/o mapping      & ×          & ×         & 0.14          & 0.35         & 0.16     & 1.23    & \textcolor{cyan}{\textbf{0.09}}     & 0.26    & 1.17     & 4.63    & 0.27      & 0.92 & 0.16 & 0.81 &0.33 &1.37    \\
			& LO-Net                  & 1.05    & 1.78    & 0.26     & 0.49    & 0.30     & 0.36    & 0.57    & 0.14   & 3.29     & 3.07    & 1.00      & 1.12 &0.77 &1.45 &1.03 &1.20     \\\cmidrule(r){1-18}
			\multirow{2}{*}{Camera-based methods} &
			DeepVO      & 1.34          & 2.75         & 1.87          & 1.46        & 1.56     & 2.01    & 0.82     & 0.48    & 5.14     & 4.04    & 1.91      & 1.98 &1.41 &2.22 &2.00 &2.13    \\
			& TartanVO    & 0.78    & 0.25    & 0.19     & 0.12    & 0.37     & 0.11    & 0.55    & 0.04   & 1.01     & \textcolor{cyan}{\textbf{0.64}}    & 0.64      & \textcolor{cyan}{\textbf{0.18}}  &0.48 &\textcolor{cyan}{\textbf{0.19}} &0.57 &\textcolor{cyan}{\textbf{0.22}}     \\\cmidrule(r){1-18}
			\multirow{2}{*}{4D Radar-based methods} &
			RaFlow      & 0.87          & 2.09         & 0.07          & 0.44         & 0.11    & \textcolor{cyan}{\textbf{0.09}}    & 0.13    & \textcolor{magenta}{\textbf{0.03}}    & 1.22     & 4.09    & 0.72      & 1.34 &0.25 &1.14 &0.48 &1.32     \\
			& CMFlow                  & \textcolor{cyan}{\textbf{0.06}}      & \textcolor{cyan}{\textbf{0.10}}     & \textcolor{cyan}{\textbf{0.05}}     & \textcolor{cyan}{\textbf{0.09}}    & \textcolor{cyan}{\textbf{0.09}}     & 0.14    & \textcolor{magenta}{\textbf{0.06}}      & \textcolor{magenta}{\textbf{0.03}}     & \textcolor{cyan}{\textbf{0.28}}      & 0.94     & \textcolor{cyan}{\textbf{0.14}}       & 0.29 &\textcolor{cyan}{\textbf{0.12}} &0.58 &\textcolor{cyan}{\textbf{0.11}}  &0.31    \\\cmidrule(r){1-18}
			\multicolumn{2}{c|}{4DRVO-Net}                    & \textcolor{magenta}{\textbf{0.02}}    & \textcolor{magenta}{\textbf{0.02}}    & \textcolor{magenta}{\textbf{0.01}}     & \textcolor{magenta}{\textbf{0.02}}    & \textcolor{magenta}{\textbf{0.03}}     & \textcolor{magenta}{\textbf{0.05}}    &0.11 & \textcolor{magenta}{\textbf{0.03}}     & \textcolor{magenta}{\textbf{0.26}}    & \textcolor{magenta}{\textbf{0.08}}     & \textcolor{magenta}{\textbf{0.06}}    & \textcolor{magenta}{\textbf{0.09}}      & \textcolor{magenta}{\textbf{0.09}}     & \textcolor{magenta}{\textbf{0.18}}& \textcolor{magenta}{\textbf{0.08}} & \textcolor{magenta}{\textbf{0.07}} \\ \bottomrule
		\end{tabular}
	}
\end{table*}

\begin{table*}[htbp]
	\renewcommand{\arraystretch}{1.2}
	\caption{The 4D radar-visual odometry and 64-line LiDAR odometry experiment results on the VoD dataset. The best result is bolded in red.}
	\label{table2}
	\centering
	\resizebox{\textwidth}{!}{
		\begin{tabular}{c|cc|cc|cc|cc|cc|cc|cc|cc}
			\toprule
			\multicolumn{1}{c|}{\multirow{2}{*}{Method}} & \multicolumn{2}{c|}{03} & \multicolumn{2}{c|}{04} & \multicolumn{2}{c|}{09} & \multicolumn{2}{c|}{17} & \multicolumn{2}{c|}{19} & \multicolumn{2}{c|}{22} & \multicolumn{2}{c|}{24} & \multicolumn{2}{c}{Mean} \\ \cmidrule(l){2-17} 
			&$t_{rel}$          & $r_{rel}$         & $t_{rel}$          & $r_{rel}$         & $t_{rel}$          & $r_{rel}$         & $t_{rel}$          & $r_{rel}$         & $t_{rel}$          & $r_{rel}$         & $t_{rel}$           & $r_{rel}$  & $t_{rel}$          & $r_{rel}$   & $t_{rel}$          & $r_{rel}$        \\ \cmidrule(r){1-17} 
			
		A-LOAM w/o mapping      & 0.06          & 0.10         & 0.03          & 0.12         & 0.06     & 0.06    & \textcolor{magenta}{\textbf{0.03}}     & \textcolor{magenta}{\textbf{0.02}}     & \textcolor{magenta}{\textbf{0.14}}      & 0.19    & \textcolor{magenta}{\textbf{0.06}}      & 0.11 & \textcolor{magenta}{\textbf{0.06}}  & \textcolor{magenta}{\textbf{0.08}} &\textcolor{magenta}{\textbf{0.06}} &0.10    \\ \cmidrule(l){1-17}
		4DRVO-Net & \textcolor{magenta}{\textbf{0.02}}    & \textcolor{magenta}{\textbf{0.02}}    & \textcolor{magenta}{\textbf{0.01}}     & \textcolor{magenta}{\textbf{0.02}}    & \textcolor{magenta}{\textbf{0.03}}     & \textcolor{magenta}{\textbf{0.05}}    &0.11 & 0.03     & 0.26    & \textcolor{magenta}{\textbf{0.08}}     & \textcolor{magenta}{\textbf{0.06}}    & \textcolor{magenta}{\textbf{0.09}}      & 0.09     & 0.18& 0.08 & \textcolor{magenta}{\textbf{0.07}} \\
		\bottomrule
		\end{tabular}
	}
\end{table*}

\subsection{Baselines}
The spatial representation of 4D radar points, unlike traditional 3D radar, is in three dimensions (X, Y, Z), making them more akin to LiDAR points. Consequently, our comparative analysis revolves around both 3D point-based odometry methods and image-based odometry methods. Specifically, we consider four classical methods: ICP-point2point \cite{besl1992method} (ICP-po2po), ICP-point2plane \cite{chen1992object} (ICP-po2pl), GICP \cite{generalized}, and NDT \cite{stoyanov2012fast}. Additionally, we examine two LiDAR-based methods: A-LOAM w/o mapping \cite{zhang2017low} and LO-Net \cite{lonet}, as well as two image-based methods: DeepVO \cite{deepvo} and TartanVO \cite{tartanvo}. Furthermore, we conduct a comparison of two 4D radar-based odometry approaches: RaFlow \cite{raflow} and CMFlow \cite{cmflow}.
\subsection{Evaluation Metrics}
The performance of the proposed method was evaluated using the relative pose error (RPE) to quantify the difference between the estimated pose and the ground-truth pose. The RPE evaluates the accuracy of the algorithm by measuring the difference between the pose variation of the estimated pose and the ground-truth pose at intervals or distances, which considers both rotation and translation errors. Furthermore, in this study, we used the average translational root mean square error (RMSE) (m/m) and average rotational RMSE (°/m) for all possible subsequences with lengths ranging from 20 to 160 m, in steps of 20 m, for comparison purposes.
\subsection{Training Details}
\label{Training Details}
Due to the distinct 4D radar equipment utilized in the VoD dataset and the In-house dataset, there is a significant disparity in the number of 4D radar points collected per frame. Therefore, for the VoD dataset we sampled the input 4D radar point cloud to $N=256$ and for the In-house dataset we sampled the input 4D radar point cloud to $N=512$. All the training and evaluation experiments were performed on a single NVIDIA 3090 GPU with Pytorch 1.8.0. The Adam optimizer is adopted. The initial learning rate was 0.001 and decayed by 0.1 after every 10 epochs. The epoch was 40, and the batchsize was 8. The initial values of trainable parameters $s_e$ and $s_t$ in Eq. \ref{equation11} were set to -2.5 and 0, respectively. In Eq. \ref{equation12}, $\lambda^1=1$, $\lambda^2=2$, $\lambda^3=4$, $\lambda^4=8$. 

\begin{figure*}[htbp]
	\centering
	\subfloat{
		\includegraphics[width=0.99\linewidth]{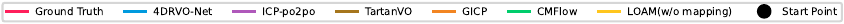}
	}
	\vspace{-3.6mm}
	\setcounter{subfigure}{0}
	\subfloat[2D Trajectory Plots of Seq.03]{
		\includegraphics[width=0.3\linewidth]{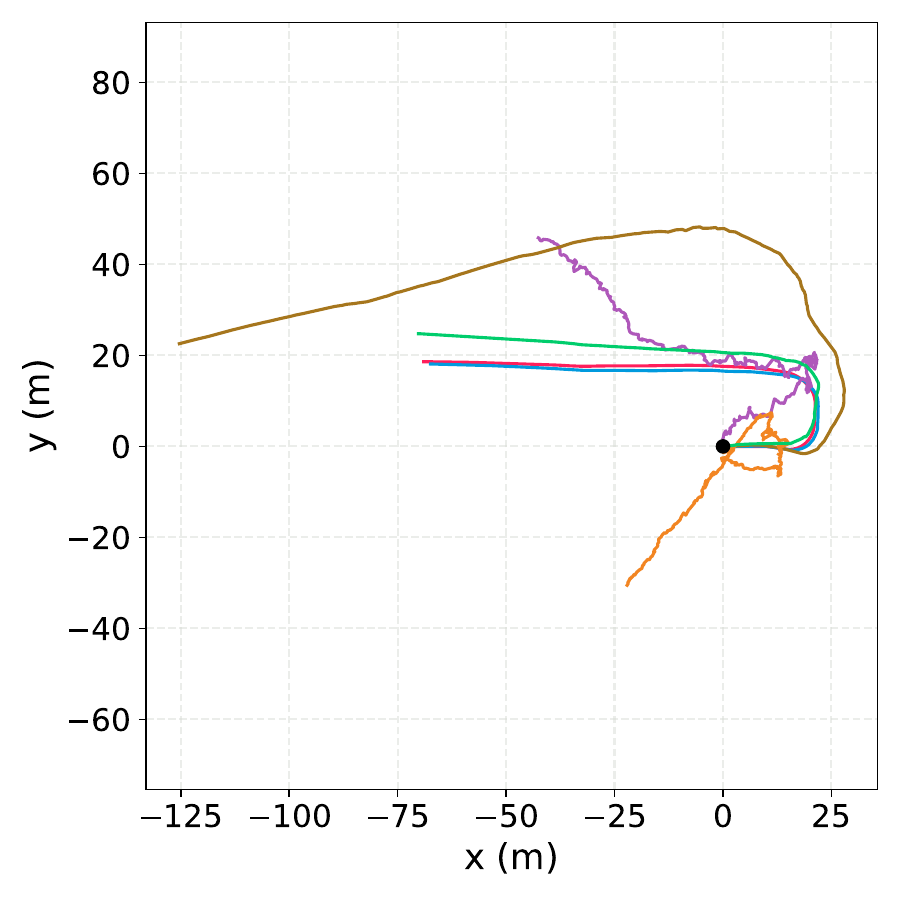}
	}
	\quad
	\subfloat[2D Trajectory Plots of Seq.04]{
		\includegraphics[width=0.3\linewidth]{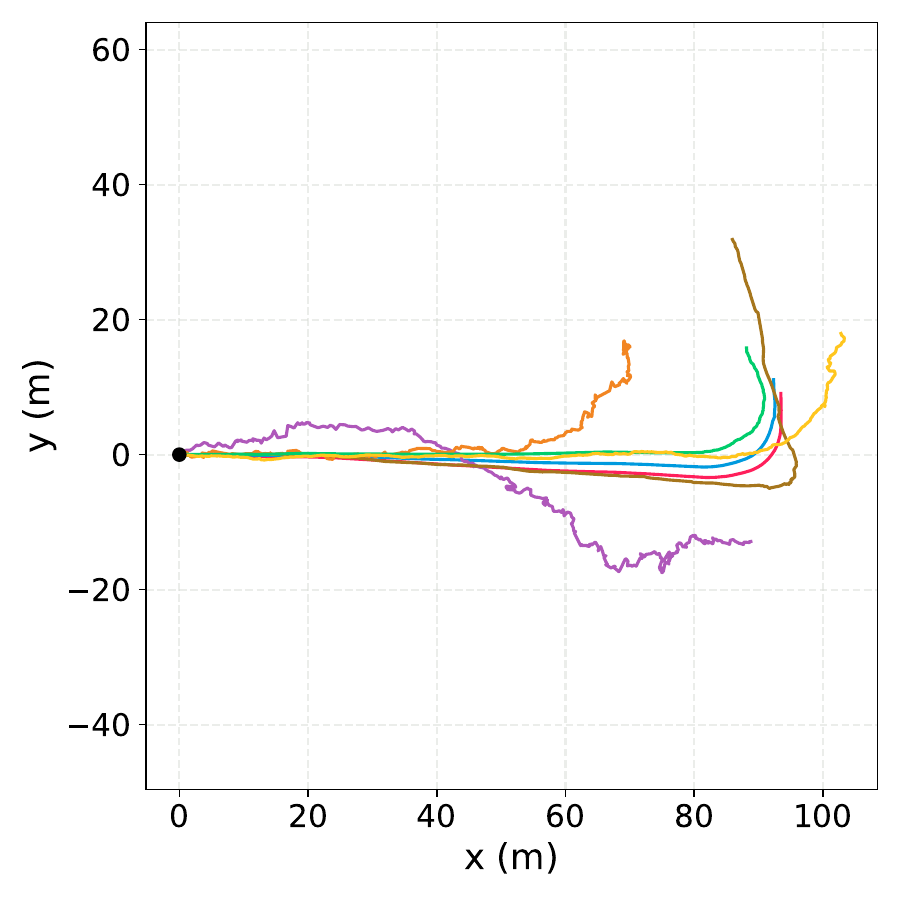}
	}
	\quad
	\subfloat[2D Trajectory Plots of Seq.09]{
		\includegraphics[width=0.3\linewidth]{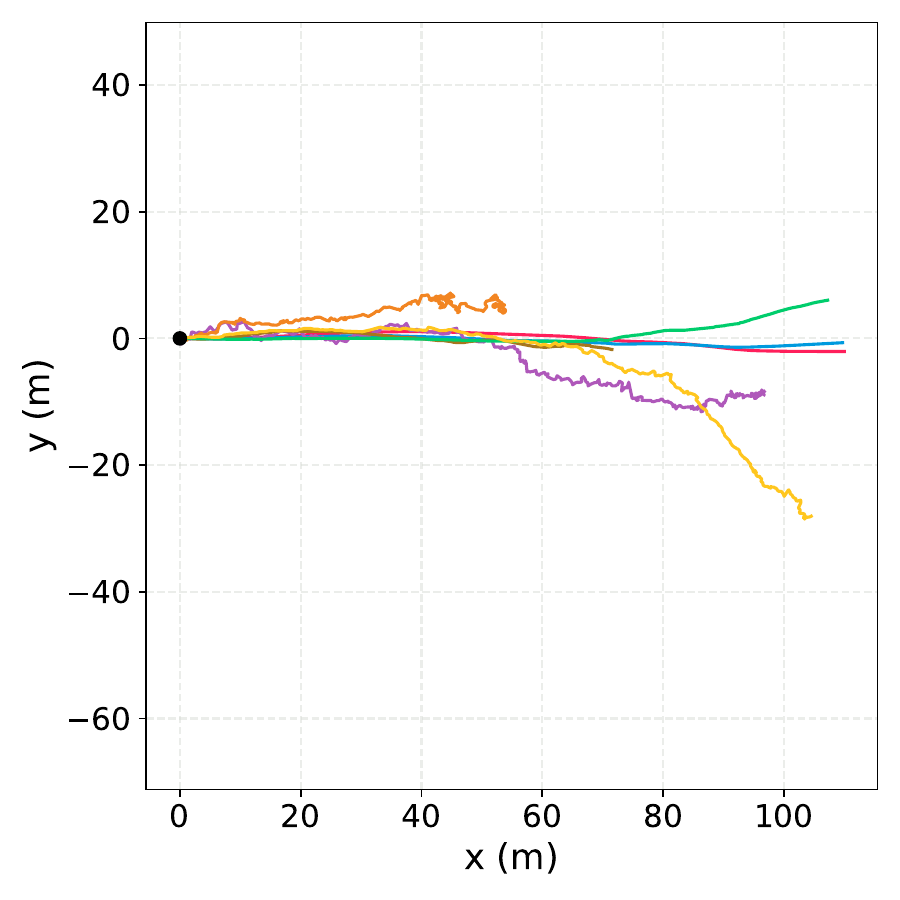}
	}
	\quad
	\subfloat[2D Trajectory Plots of Seq.17]{
		\includegraphics[width=0.3\linewidth]{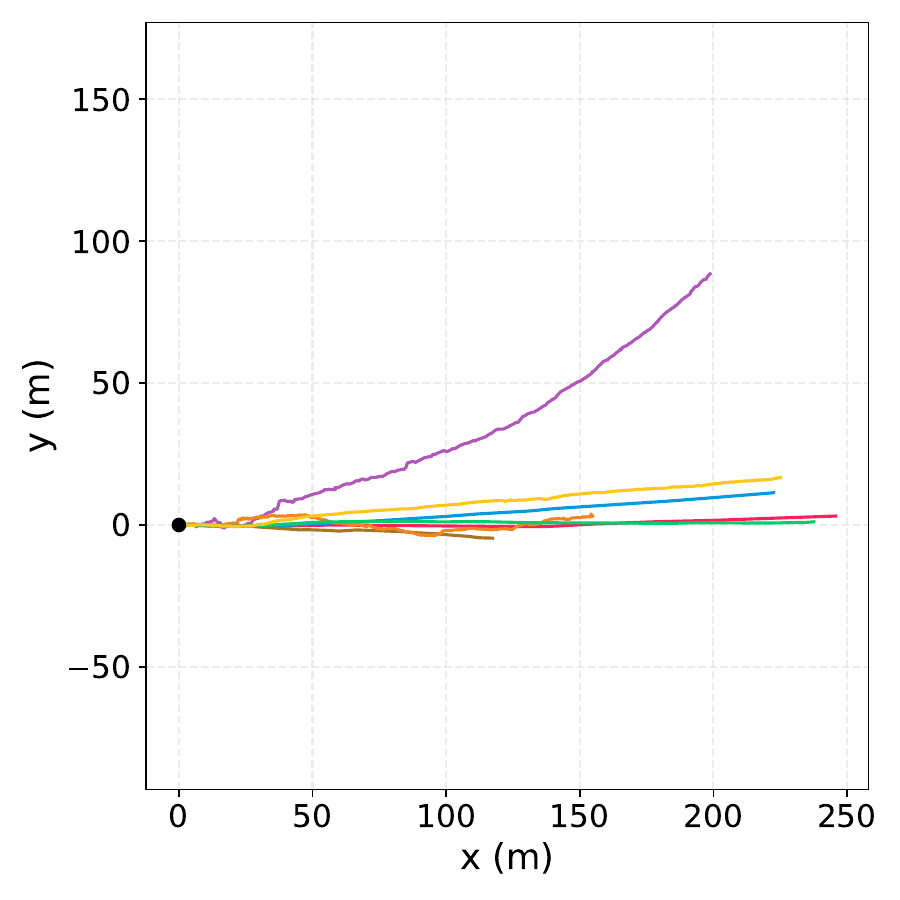}
	}
	\quad
	\subfloat[2D Trajectory Plots of Seq.22]{
		\includegraphics[width=0.3\linewidth]{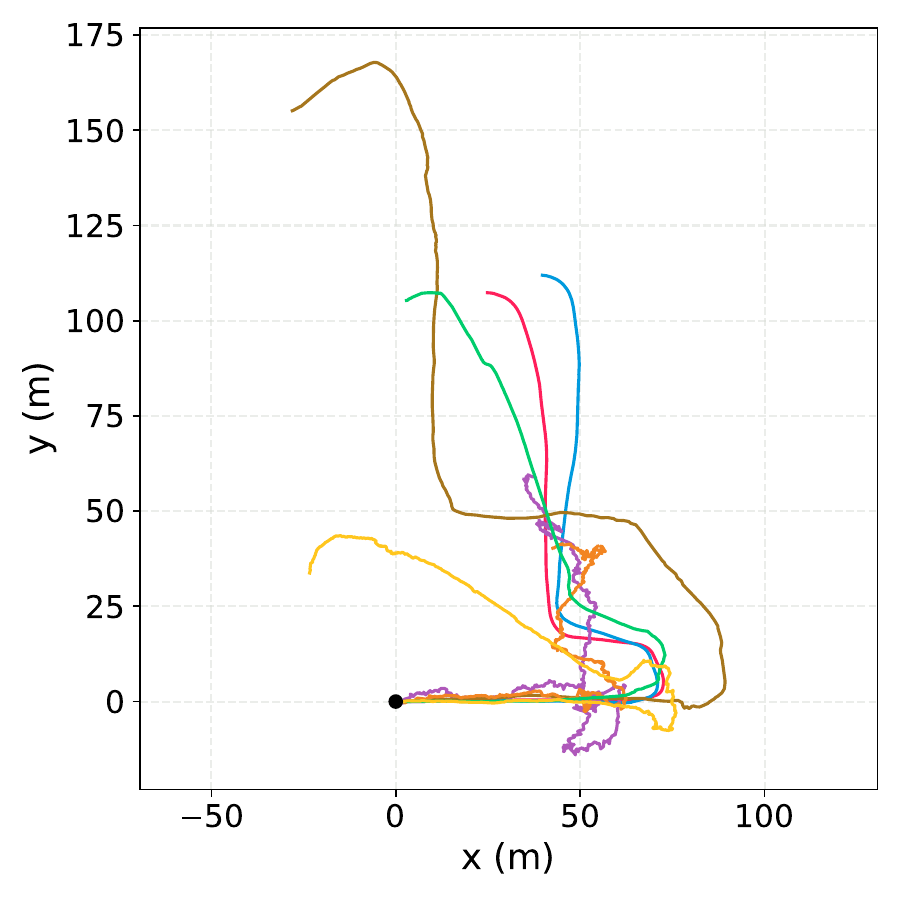}
	}
	\quad
	\subfloat[2D Trajectory Plots of Seq.24]{
		\includegraphics[width=0.3\linewidth]{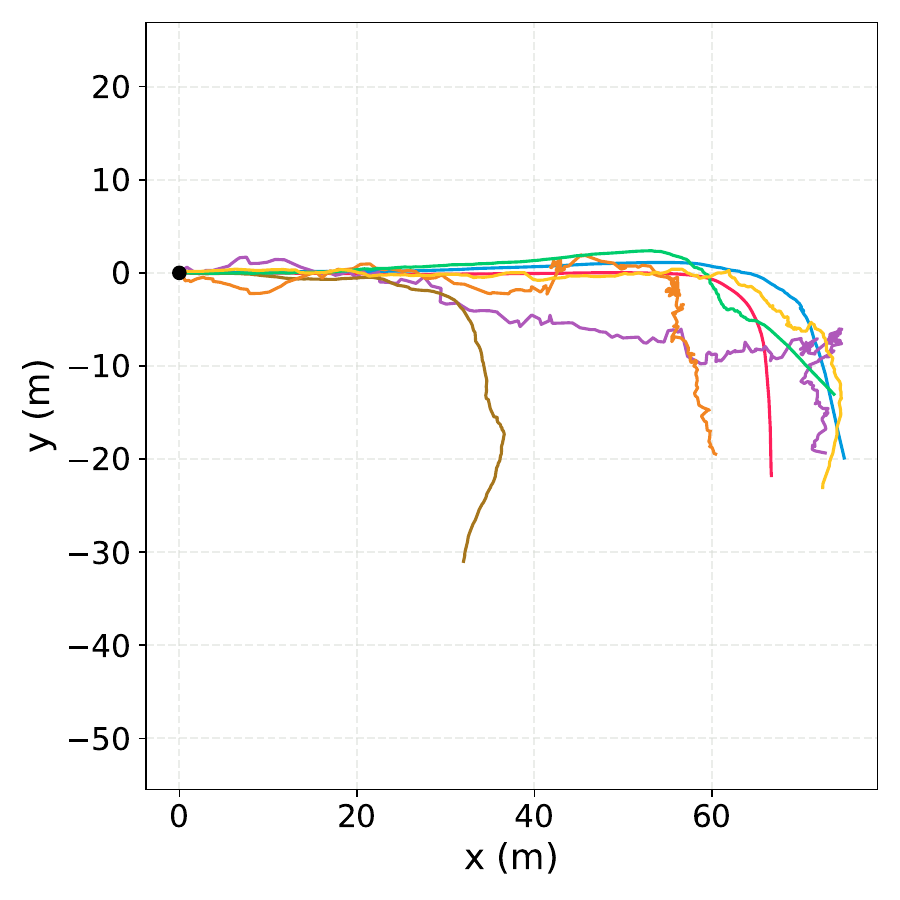}
	}
	\caption{Trajectory results of the proposed method, ICP-po2po, TartanVO, GICP, and CMFlow on sequences 03, 04, 09, 17, 22 and 24. Our algorithms demonstrate significant improvements in terms of both algorithmic robustness and localization accuracy compared to the other methods. \label{fig6}}
\end{figure*}
\begin{figure}[]
	\centering
	\includegraphics[width=0.98\linewidth]{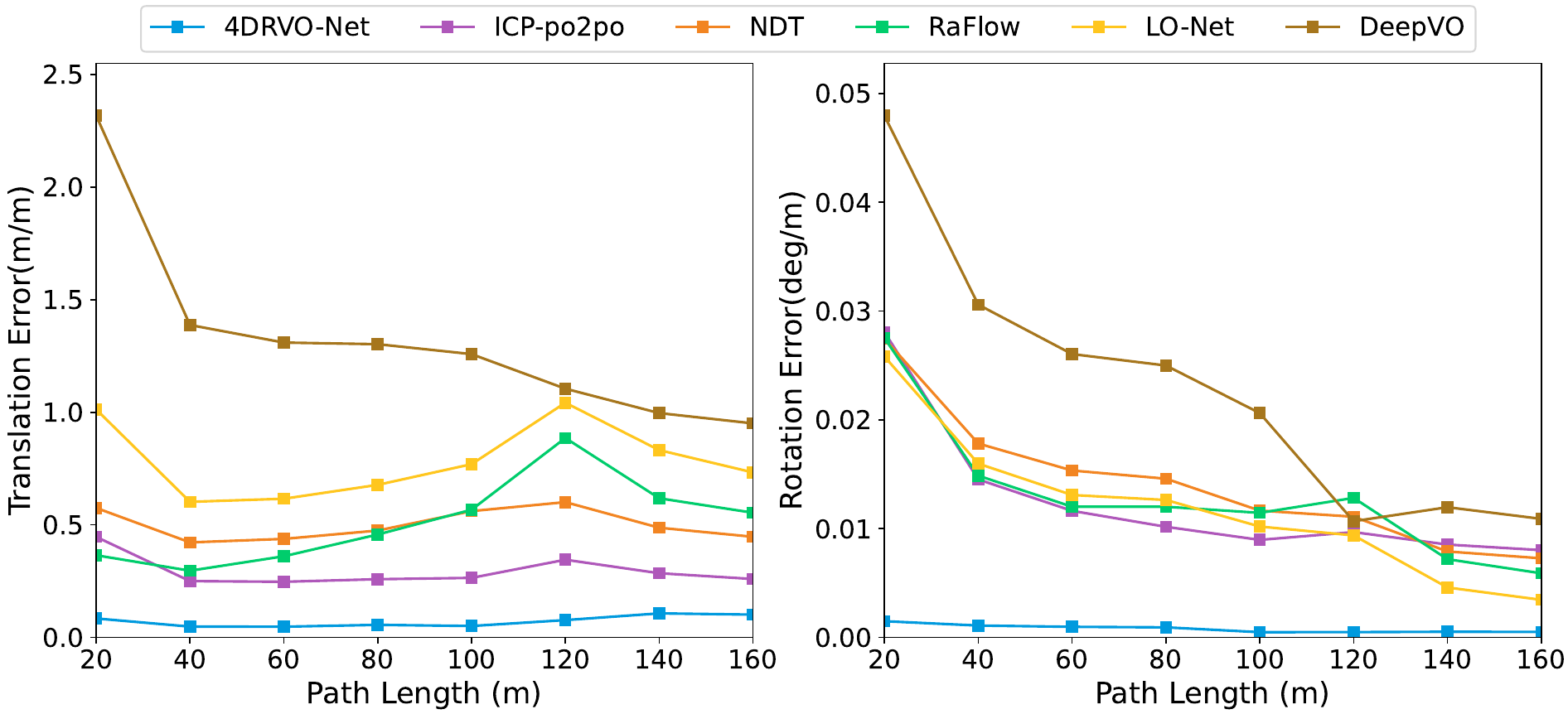}
	\caption{Average translational and rotational error on the VoD test sequences on all possible subsequences for path lengths of $20, 40, ..., 160$ m. Our method has the best performance.}
	\label{fig7}
\end{figure}
\section{Experimental Results}
\label{section5}
In this section, the quantitative and qualitative results of the network performance for the 4D radar odometry task are presented. Subsequently, the results of an extensive ablation study conducted to demonstrate the effect of each component on the results are reported. Finally, point confidence is visualized and discussed.
\subsection{Performance Evaluation}
\subsubsection{Results on the VoD Dataset}

Table \ref{table1} presents the quantitative comparison results of the performance of the proposed method on a test set of the VoD dataset compared with the baseline methods. Our method exhibited superior performance compared with other methods in terms of both the relative translation error and relative rotation error metrics on the VoD dataset. Specifically, our method surpassed the second-best method by $27.5\%$ in mean relative translation error and $69.3\%$ in mean relative rotation error. Despite their satisfactory performance on dense point clouds (e.g., LiDAR), the sparsity and noisiness of radar data make it challenging for state-of-the-art methods to provide the same efficacy on radar data. Furthermore, the A-LOAM-based method was unable to operate completely on the 03 sequence owing to the sparsity of the 4D radar point cloud. The scenes in the VoD dataset consisted of numerous dynamic objects, which significantly impaired the performance of the camera-based odometry method, rendering it non-robust for the VoD dataset. The proposed method leverages the capabilities of 4D radar in dynamic scenes, incorporates abundant semantic information from images, and can be trained to iteratively refine the estimated poses multiple times in a single network inference. Consequently, 4D radar visual odometry demonstrates outstanding performance in complex scenes.

In addition, we also compared the results of the proposed 4D radar-visual odometry method with the 64-line LiDAR odometry results of A-LOAM without mapping optimization. As shown in Table \ref{table2}, the 4DRVO-Net using the low-cost 4D radar and monocular camera achieved a comparable localization effect to the 64-line LiDAR odometry. This depends on the effective design of each module in the proposed method.

The qualitative results are shown in Figs. \ref{fig6} and \ref{fig7}. Fig. \ref{fig6} shows the XY plane projection results of the proposed method, ICP-po2po, GICP, TartanVO, and CMFlow trajectories on sequences 03, 04, 09, 17, 22, and 24 of the VoD datasets. Owing to the spatial unevenness and noise in 4D radar point clouds, the trajectories generated by classical odometry methods exhibit substantial deviations from the ground truth trajectory. Furthermore, camera-based odometry methods lack direct access to absolute scales, resulting in generated trajectories that do not align with the ground-truth trajectory scale. 4D radar captures precise spatial information about the environment, enabling odometry estimates at absolute scales, and generating trajectories that closely align with the ground truth trajectory.
\begin{table*}[htbp]
	\renewcommand{\arraystretch}{1.2}
	\caption{The 4D radar-visual odometry experiment results on the in-house dataset. The best result is bolded in red, the second best is bolded in blue.}
	\label{table3}
	\centering
	\begin{tabular}{cc|cc|cc|cc}
		\toprule
		\multicolumn{2}{c}{\multirow{2}{*}{Method}} & \multicolumn{2}{c|}{05} & \multicolumn{2}{c|}{06} & \multicolumn{2}{c}{Mean} \\ \cmidrule(l){3-8} 
		& & $t_{rel}$          & $r_{rel}$         & $t_{rel}$          & $r_{rel}$         & $t_{rel}$          & $r_{rel}$          \\ \cmidrule(r){1-8} 
		\multirow{4}{*}{Classical-based methods} & ICP-po2po               & 0.39                & 0.83                & 0.50                & 0.81                & 0.45               & 0.82     \\
		& ICP-po2pl               & 0.43                & 1.20                & 0.50                & 1.43                & 0.47     & 1.32\\
		& GICP                    & 0.22                & \textcolor{cyan}{\textbf{0.21}}                & 0.25                & 0.42                & 0.24    & \textcolor{cyan}{\textbf{0.32}} \\
		& NDT                     & \textcolor{cyan}{\textbf{0.09}}                 & 0.42                & \textcolor{cyan}{\textbf{0.11}}                & \textcolor{cyan}{\textbf{0.26}}                & \textcolor{cyan}{\textbf{0.10}}    & 0.34\\
		\cmidrule(r){1-8}
		\multirow{2}{*}{LiDAR-based methods} &
		A-LOAM w/o mapping      & 0.36                & 1.07                & 0.19                & 0.72                & 0.28    & 0.90 \\
		& LO-Net                  & 0.30                & 0.51                 & 0.41                & 0.60                 & 0.36    & 0.56 \\
		\cmidrule(r){1-8}
		\multirow{1}{*}{Camera-based methods} &
		TartanVO      & 0.85                & 1.05                & 1.37                & 1.26                & 1.11    & 1.16 \\
		\cmidrule(r){1-8}
		\multirow{1}{*}{4D Radar-based methods}
		& RaFlow                  & 0.29                & 0.34                 & 0.31                & 0.41                 & 0.30    & 0.38 \\
		\cmidrule(r){1-8}
		\multicolumn{2}{c|}{4DRVO-Net}  & \textcolor{magenta}{\textbf{0.03}}                & \textcolor{magenta}{\textbf{0.06}}                & \textcolor{magenta}{\textbf{0.06}}                & \textcolor{magenta}{\textbf{0.07}}                & \textcolor{magenta}{\textbf{0.05}}  & \textcolor{magenta}{\textbf{0.07}} \\ \bottomrule
	\end{tabular}
\end{table*}
\begin{figure}[tb]
	\centering
	\subfloat{
		\includegraphics[width=0.99\linewidth]{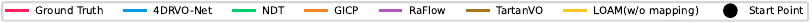}
	}
	\vspace{-3.6mm}
	\setcounter{subfigure}{0}
	\subfloat[2D Trajectory Plots of Seq.05]{
		\includegraphics[width=0.45\linewidth]{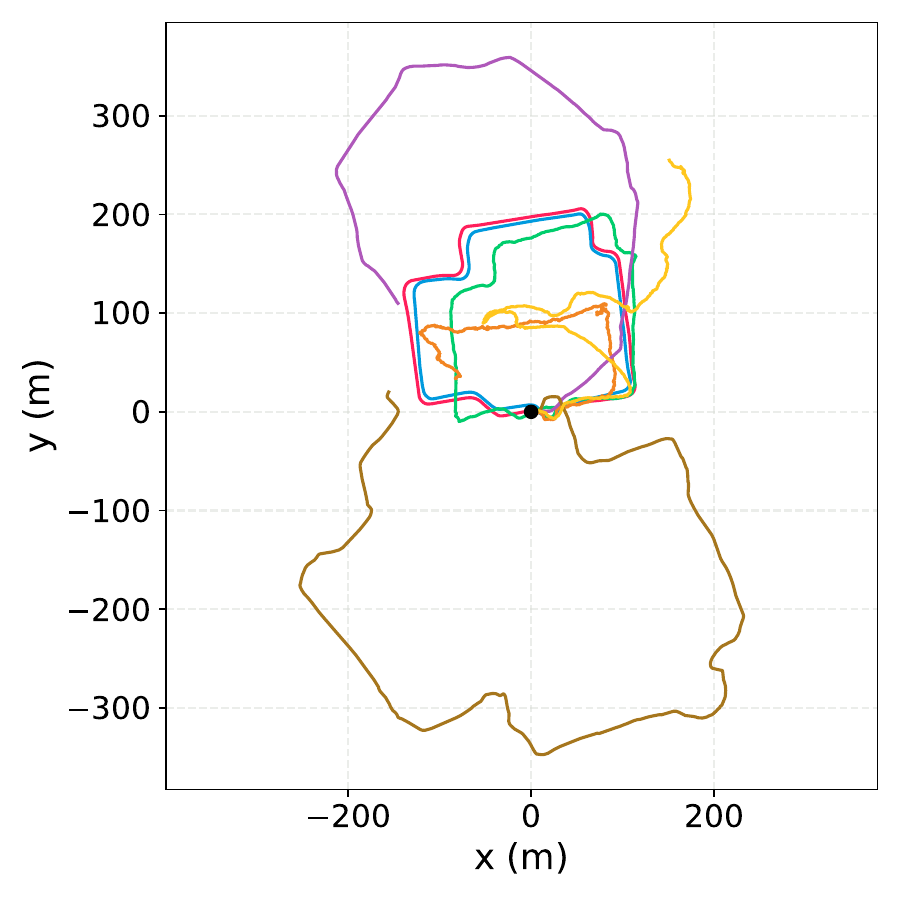}
	}
	\quad
	\subfloat[3D Trajectory Plots of Seq.05]{
		\includegraphics[width=0.45\linewidth]{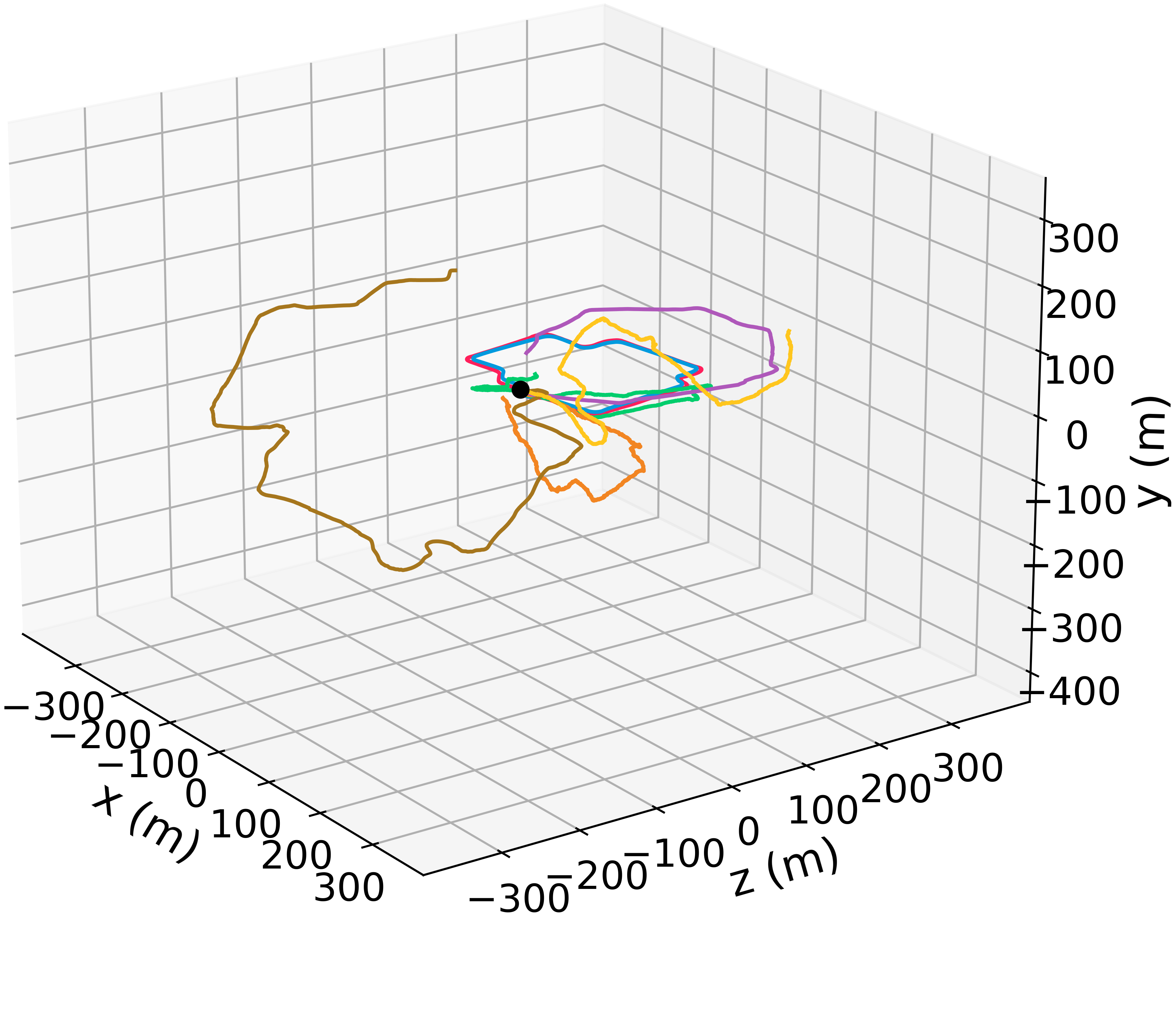}
	}
	\caption{The 2D and 3D trajectories of the proposed algorithm, GICP, NDT, LOAM, TartanVO, and RaFlow are depicted for the test sequence 05 of the in-house dataset. Our method obtains the most accurate trajectory. \label{fig8}}
\end{figure}

\begin{figure}[tb]
	\centering
	\includegraphics[width=0.98\linewidth]{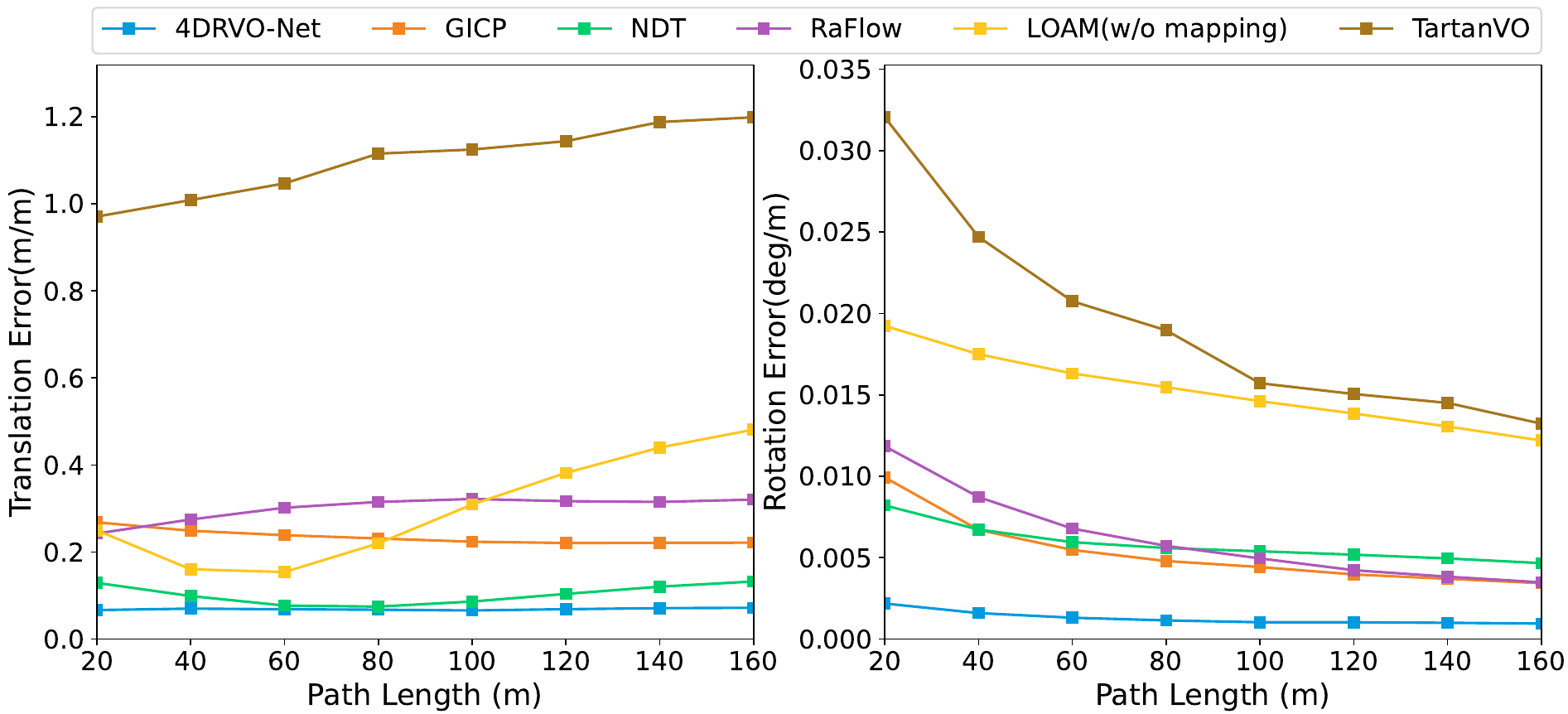}
	\caption{Average translational and rotational error on the in-house dataset test sequences on all possible subsequences for path lengths of $20, 40, ..., 160$ m. Our method has the best performance.}
	\label{fig9}
\end{figure}
\subsubsection{Results on the In-house Dataset}
Table \ref{table3} presents the evaluation results for the in-house dataset. The proposed 4D radar--visual odometry method achieved optimal localization performance for both test sequences, with exceptional performance in terms of relative rotation error. Specifically, our method surpassed the second-best method by $35\%$ in terms of the average relative translation error and $76.2\%$ in terms of the average relative rotation error. Furthermore, we observed a substantial improvement in the localization performance of classical-based odometry methods (e.g., NDT and GICP) on the in-house dataset compared with the VoD dataset. This disparity can be attributed to the fact that the number of 4D radar points in a single frame of the point cloud from our dataset was twice that of the point cloud from the VoD dataset because of the differences in the 4D radar device used. Consequently, the performance of the point-cloud-alignment-based odometry method was improved.

The qualitative results are shown in Figs. \ref{fig8} and \ref{fig9}. Fig.  \ref{fig8} shows the 2D and 3D trajectories of the proposed method and the comparison method on test sequence 05. The figure demonstrates that the proposed method maintains an impressive localization performance on long-distance datasets and exhibits minimal trajectory drift along the X-, Y-, and Z-axis compared with the other algorithms. Fig. \ref{fig9} illustrates the average translational and rotational error of our method, GICP, NDT, RaFlow, LOAM (w/o mapping), and TartanVO across all test sequences, considering all possible subsequences with lengths ranging from 20 to 160 m in 20-m steps. The figure clearly demonstrates that the proposed method achieves superior positioning performance across all sub-sequences.

\subsection{Ablation Study}
To analyze the effectiveness of each module in the proposed method, an ablation study was conducted on the VoD dataset by removing or changing the module components. The training and testing conditions were identical to those described in Sec. \ref{Training Details}.

\textbf{Benefits of Radar-PointNet++:}
The feature aggregation method in the Radar-PointNet++ module was modified to input the spatial, velocity, and intensity information of 4D radar points into a unified MLP for feature extraction. This means that all feature information of 4D radar points was mapped to the same feature space. The model exhibited a significant decrease in localization performance, as indicated by Table \ref{table4}\textcolor{blue}{-a}. This is because the proposed feature aggregation method conducts multi-scale feature extraction and deep fusion among various feature information, thereby maximizing the utilization of the feature information from 4D radar points. Consequently, the proposed feature-aggregation method achieved superior results.
\begin{table*}[htbp]
	\renewcommand{\arraystretch}{1.2}
	\caption{Results of the ablation study. We report performance comparisons between the full model and the model with components of the module removed or changed on the VoD dataset.}
	\label{table4}
	\centering
	\resizebox{\textwidth}{!}{
		\begin{tabular}{cc|cc|cc|cc|cc|cc|cc|cc|cc}
			\toprule
			\multicolumn{2}{c|}{\multirow{2}{*}{Method}} & \multicolumn{2}{c|}{03} & \multicolumn{2}{c|}{04} & \multicolumn{2}{c|}{09} & \multicolumn{2}{c|}{17} & \multicolumn{2}{c|}{19} & \multicolumn{2}{c|}{22} & \multicolumn{2}{c|}{24} & \multicolumn{2}{c}{Mean} \\ \cmidrule(l){3-18} 
			& & $t_{rel}$          & $r_{rel}$         & $t_{rel}$          & $r_{rel}$         & $t_{rel}$          & $r_{rel}$         & $t_{rel}$          & $r_{rel}$         & $t_{rel}$          & $r_{rel}$         & $t_{rel}$           & $r_{rel}$   & $t_{rel}$          & $r_{rel}$	& $t_{rel}$          & $r_{rel}$       \\ \cmidrule(r){1-18} 
			\multirow{2}{*}{(a)} &Ours(w/o Radar-PointNet++) &0.06 & 0.07 & 0.03 & \textcolor{magenta}{\textbf{0.02}} & 0.11 & 0.07 & \textcolor{magenta}{\textbf{0.10}} & 0.04 & 0.48 & 0.32 & 0.07 & 0.10 &0.17 &0.41 &0.15 &0.15 \\
			&Ours(Full, with Radar-PointNet++) &\textcolor{magenta}{\textbf{0.02}}    & \textcolor{magenta}{\textbf{0.02}}    & \textcolor{magenta}{\textbf{0.01}}     & \textcolor{magenta}{\textbf{0.02}}    & \textcolor{magenta}{\textbf{0.03}}     & \textcolor{magenta}{\textbf{0.05}}    & 0.11    & \textcolor{magenta}{\textbf{0.03}}    & \textcolor{magenta}{\textbf{0.26}}     & \textcolor{magenta}{\textbf{0.08}}    & \textcolor{magenta}{\textbf{0.06}}      & \textcolor{magenta}{\textbf{0.09}} & \textcolor{magenta}{\textbf{0.09}} & \textcolor{magenta}{\textbf{0.18}} & \textcolor{magenta}{\textbf{0.08}} & \textcolor{magenta}{\textbf{0.07}}\\
			\cmidrule(r){1-18} 
			
			\multirow{2}{*}{(b)} &Ours(w/o Point Confidence) &0.04 & 0.04 & \textcolor{magenta}{\textbf{0.01}} & 0.04 & 0.04 & 0.06 & 0.15 & \textcolor{magenta}{\textbf{0.02}} & 0.38 & 0.23 & \textcolor{magenta}{\textbf{0.05}} & \textcolor{magenta}{\textbf{0.09}} &0.10 &\textcolor{magenta}{\textbf{0.18}} &0.11 &0.09\\
			&Ours(Full, with Point Confidence) &\textcolor{magenta}{\textbf{0.02}}    & \textcolor{magenta}{\textbf{0.02}}    & \textcolor{magenta}{\textbf{0.01}}     & \textcolor{magenta}{\textbf{0.02}}    & \textcolor{magenta}{\textbf{0.03}}     & \textcolor{magenta}{\textbf{0.05}}    & \textcolor{magenta}{\textbf{0.11}}    & 0.03    & \textcolor{magenta}{\textbf{0.26}}     & \textcolor{magenta}{\textbf{0.08}}    & 0.06      & \textcolor{magenta}{\textbf{0.09}} & \textcolor{magenta}{\textbf{0.09}} & \textcolor{magenta}{\textbf{0.18}} & \textcolor{magenta}{\textbf{0.08}} & \textcolor{magenta}{\textbf{0.07}}\\
			\cmidrule(r){1-18} 
			
			\multirow{2}{*}{(c)} &Ours(Feature Concatenation) &0.05 & 0.07 & 0.05 & 0.06 & 0.06 & 0.06 & 0.23 & 0.10 & 0.68 & \textcolor{magenta}{\textbf{0.07}} & 0.09 & 0.13 & 0.27 & 0.77 & 0.20 & 0.18\\
			&Ours(Transformer Attention) &0.04 & 0.07 & \textcolor{magenta}{\textbf{0.01}} & 0.04 & 0.04 & \textcolor{magenta}{\textbf{0.04}} & \textcolor{magenta}{\textbf{0.06}} & \textcolor{magenta}{\textbf{0.01}} & 0.29 & 0.22 & 0.07 & 0.10 & 0.17 & 0.40 & 0.10 & 0.13\\
			&Ours(Full, with Adaptive Feature Fusion)&\textcolor{magenta}{\textbf{0.02}}    & \textcolor{magenta}{\textbf{0.02}}    & \textcolor{magenta}{\textbf{0.01}}     & \textcolor{magenta}{\textbf{0.02}}    & \textcolor{magenta}{\textbf{0.03}}     &0.05    & 0.11   & 0.03    & \textcolor{magenta}{\textbf{0.26}}     & 0.08    & \textcolor{magenta}{\textbf{0.06}}      & \textcolor{magenta}{\textbf{0.09}} & \textcolor{magenta}{\textbf{0.09}} & \textcolor{magenta}{\textbf{0.18}} & \textcolor{magenta}{\textbf{0.08}} & \textcolor{magenta}{\textbf{0.07}}\\
			\cmidrule(r){1-18} 
			
			\multirow{2}{*}{(d)} &Ours(w/o Pose Warp Refinement) &0.49 & 0.85 & 0.18 & 0.29 & 0.27 & 0.33 & 0.42 & 0.16 & 1.93 & 0.33 & 0.37 & 0.64 & 0.56 & 1.67 & 0.60 & 0.61\\
			&Ours(with Pose Warp Refinement, $l=1$) &0.26 & 0.45 & 0.19 & 0.15 & 0.19 & 0.11 & 0.41 & 0.11 & 1.78 & 0.15 & 0.27 & 0.37 & 0.43 & 0.61 &0.50 &0.28\\
			&Ours(w/o Pose Warp Refinement, $l=2$) &0.05 & 0.07 & 0.04 & 0.05 & 0.08 & 0.09 & 0.20 & 0.05 & 0.99 & 0.28 & 0.11 & 0.15 & 0.21 & 0.40 & 0.24 & 0.16\\
			&Ours(Full, with Pose Warp Refinement, $l=3$)&\textcolor{magenta}{\textbf{0.02}}    & \textcolor{magenta}{\textbf{0.02}}    & \textcolor{magenta}{\textbf{0.01}}     & \textcolor{magenta}{\textbf{0.02}}    & \textcolor{magenta}{\textbf{0.03}}     & \textcolor{magenta}{\textbf{0.05}}    & \textcolor{magenta}{\textbf{0.11}}    & \textcolor{magenta}{\textbf{0.03}}    & \textcolor{magenta}{\textbf{0.26}}     & \textcolor{magenta}{\textbf{0.08}}    & \textcolor{magenta}{\textbf{0.06}}      & \textcolor{magenta}{\textbf{0.09}} & \textcolor{magenta}{\textbf{0.09}} & \textcolor{magenta}{\textbf{0.18}} & \textcolor{magenta}{\textbf{0.08}} & \textcolor{magenta}{\textbf{0.07}}\\
			\cmidrule(r){1-18} 
			\multirow{2}{*}{(e)} &Ours(w/o Point Confidence and Embedding Feature Refinement) &0.06 & \textcolor{magenta}{\textbf{0.02}} & 0.03 & \textcolor{magenta}{\textbf{0.02}} & 0.07 & 0.07 & \textcolor{magenta}{\textbf{0.10}} & \textcolor{magenta}{\textbf{0.01}} &\textcolor{magenta}{\textbf{0.18}} & 0.11 & 0.11 & 0.14 &\textcolor{magenta}{\textbf{0.09}} &\textcolor{magenta}{\textbf{0.12}} &0.09 &\textcolor{magenta}{\textbf{0.07}}\\
			&Ours(Full, with Point Confidence and Embedding Feature Refinement) & \textcolor{magenta}{\textbf{0.02}}    & \textcolor{magenta}{\textbf{0.02}}    & \textcolor{magenta}{\textbf{0.01}}     & \textcolor{magenta}{\textbf{0.02}}    & \textcolor{magenta}{\textbf{0.03}}     & \textcolor{magenta}{\textbf{0.05}}    & 0.11   & 0.03    & 0.26     & \textcolor{magenta}{\textbf{0.08}}    & \textcolor{magenta}{\textbf{0.06}}      & \textcolor{magenta}{\textbf{0.09}} & \textcolor{magenta}{\textbf{0.09}} & 0.18 & \textcolor{magenta}{\textbf{0.08}} & \textcolor{magenta}{\textbf{0.07}}\\
			\bottomrule
	\end{tabular}}
\end{table*}

\textbf{Benefits of Point Confidence:}
We first removed the point-confidence estimation module, which means that dynamic and static objects in the scene contributed equally to odometry estimation. The results in Table \ref{table4}\textcolor{blue}{-b} indicate that the proposed point-confidence estimation module achieved better results, particularly in scenes with more dynamic objects, such as sequences 03, where its performance was more prominent.

\textbf{Effect of Adaptive Feature Fusion:}
In this study, we replaced the proposed A-RCFM with other fusion methods. First, we connected the image features obtained by indexing the 4D radar points on the image with the 4D radar point features as fused features, indicating that there was no interaction between the 4D radar point features and image features. Next, we employed a Transformer Attention structure to facilitate a simple interaction between the 4D radar point and image features, indicating that the 4D radar points were not adaptively selected as fusion targets of interest. Compared with our proposed adaptive fusion method, the results of both fusion methods were degraded, as displayed in Table \ref{table4}\textcolor{blue}{-c}. This is because the proposed fusion method can dynamically identify multiple image features of interest to achieve deep fusion between 4D radar point features and image features. Therefore, our proposed adaptive 4D radar--camera fusion method achieved superior results. 

\begin{figure}[tb]
	\centering
	\subfloat{
		\includegraphics[width=0.99\linewidth]{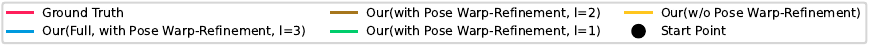}
	}
	\vspace{-3.6mm}
	\setcounter{subfigure}{0}
	\subfloat[2D Trajectory Plots of Seq.22]{
		\includegraphics[width=0.45\linewidth]{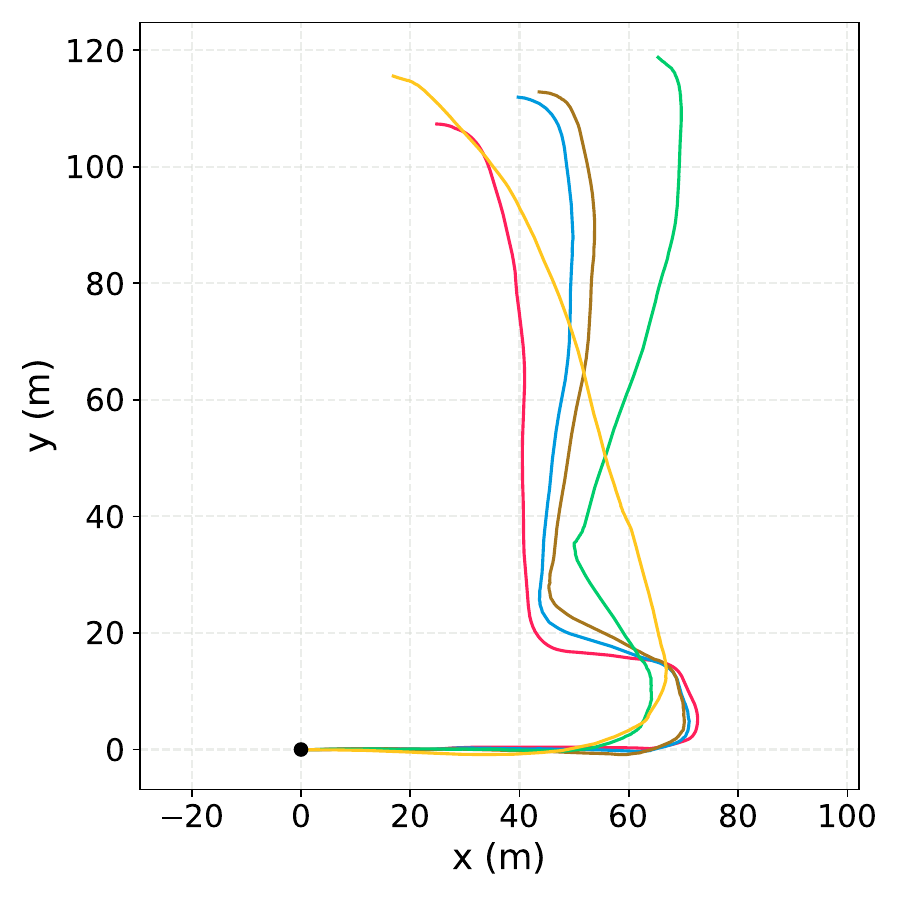}
	}
	\quad
	\subfloat[3D Trajectory Plots of Seq.24]{
		\includegraphics[width=0.45\linewidth]{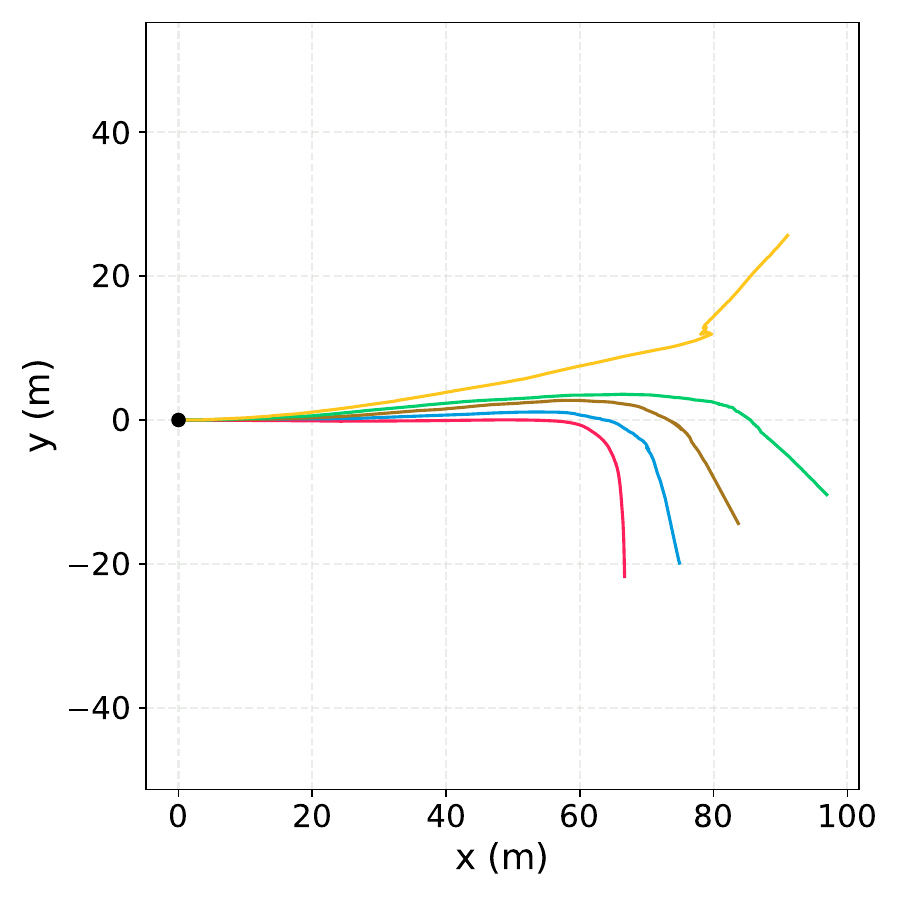}
	}
	\caption{Network architectures with complete pose warp refinement, networks with two layers of pose warp refinement, networks with one layer of pose warp refinement, and networks without pose warp refinement are evaluated on VoD dataset sequences 22 and 24. With an increasing number of pose warp refinement layers, the network's estimated poses become more accurate.\label{fig10}}
\end{figure}

\textbf{Effect of Pose Warp Refinement:}
We compared the pose estimation results of networks without pose warp refinement, with one layer of pose warp refinement, and with two layers of pose warp refinement. As indicated in Table \ref{table4}\textcolor{blue}{-d}, it is evident that with an increasing number of layers of pose warp refinement, the network achieved more accurate pose estimates, resulting in the best localization results across the full network structure with all layers of pose warp refinement, as depicted in the qualitative evaluation shown in Fig. \ref{fig10}. This highlights the significance of the coarse-to-fine refinement operations.

\textbf{Effect of Point Confidence and Embedding Feature Refinement:}
Refinement of the point confidence and embedding features was removed, indicating that these features were independently estimated at each level. Table \ref{table4}\textcolor{blue}{-e} presents the results, demonstrating that the proposed iterative refinement of the point confidence and embedding features helped to obtain better results.

\textbf{Impact of the Number of Sampling Points:}
We compared the impact of different numbers of sampled points on model performance on the In-house dataset. We sampled the input radar points to $N=256$ and $N=512$ points, respectively. As shown in Table \ref{table5}, it can be observed from the table that as the number of sampled points increases, the localization performance of the model steadily improves. Furthermore, even with a reduction in the number of sampled points leading to a decrease in model performance, the localization effect remains remarkably excellent. This demonstrates the robustness of the proposed model to varying numbers of radar points.
\begin{table*}[htbp]
	\renewcommand{\arraystretch}{1.2}
	\caption{Average runtime on the VoD dataset Seq.04}
	\label{table6}
	\centering
	\resizebox{\textwidth}{!}{
		\begin{tabular}{c|cccc|cc|cc|cc|c}
			\toprule
			\multirow{2}{*}{Method} & \multicolumn{4}{c|}{Classical method based} &\multicolumn{2}{c|}{LiDAR method based} &\multicolumn{2}{c|}{Camera method based} &\multicolumn{2}{c|}{4D Radar method based} & \multirow{2}{*}{4DRVO-Net}    \\  \cmidrule(r){2-11}
			&ICP-po2po &ICP-po2pl &GICP &NDT &A-LOAM w/o mapping &LO-Net &DeepVO &TartanVO &RaFlow &CMFlow  \\ \cmidrule(r){1-12}
			Runtime &3.80ms &1.11ms &1.29ms &1.02ms &4.70ms &11.6ms &247ms &41.6ms &36.3ms &30.4ms &39ms\\
			\bottomrule
	\end{tabular}}
\end{table*}
\begin{table}[htbp]
	\renewcommand{\arraystretch}{1.2}
	\caption{Results of the ablation study. We report the effect of different numbers of sampling points on model performance.}
	\label{table5}
	\centering
	\begin{tabular}{c|cc|cc|cc}
		\toprule
		\multirow{2}{*}{Method} & \multicolumn{2}{c|}{05} & \multicolumn{2}{c|}{06} &  \multicolumn{2}{c}{Mean} \\ \cmidrule(l){2-7} 
		& $t_{rel}$          & $r_{rel}$         & $t_{rel}$          & $r_{rel}$         & $t_{rel}$ & $r_{rel}$    \\ \cmidrule(r){1-7} 
		Ours($N=256$) &0.06 & 0.06 & 0.08 & 0.07 & 0.07 & 0.07 \\
		Ours($N=512$) &0.03 & 0.06 & 0.06 & 0.07 & 0.05 & 0.07 \\
		\bottomrule
	\end{tabular}
\end{table}
\subsection{Computation Time}
4D radar point clouds and images were sequentially captured individually over a period of time, and the timely processing of these data is crucial for maintaining the real-time nature of the algorithm. The results in Table \ref{table6} display the average runtimes of the proposed method and compare the methods for VoD dataset sequence 04. Owing to the limited number of 4D radar points in each frame, the runtime of the classical methods was significantly shorter. Our method had an overall runtime of 39 ms, which is equivalent to an approximate rate of 26 Hz. Thus, the proposed method enables real-time 4D radar--visual odometry at 26 Hz.
\subsection{Visualization of Point Confidence}
\begin{figure}[tb]
	\centering
	\includegraphics[width=0.98\linewidth]{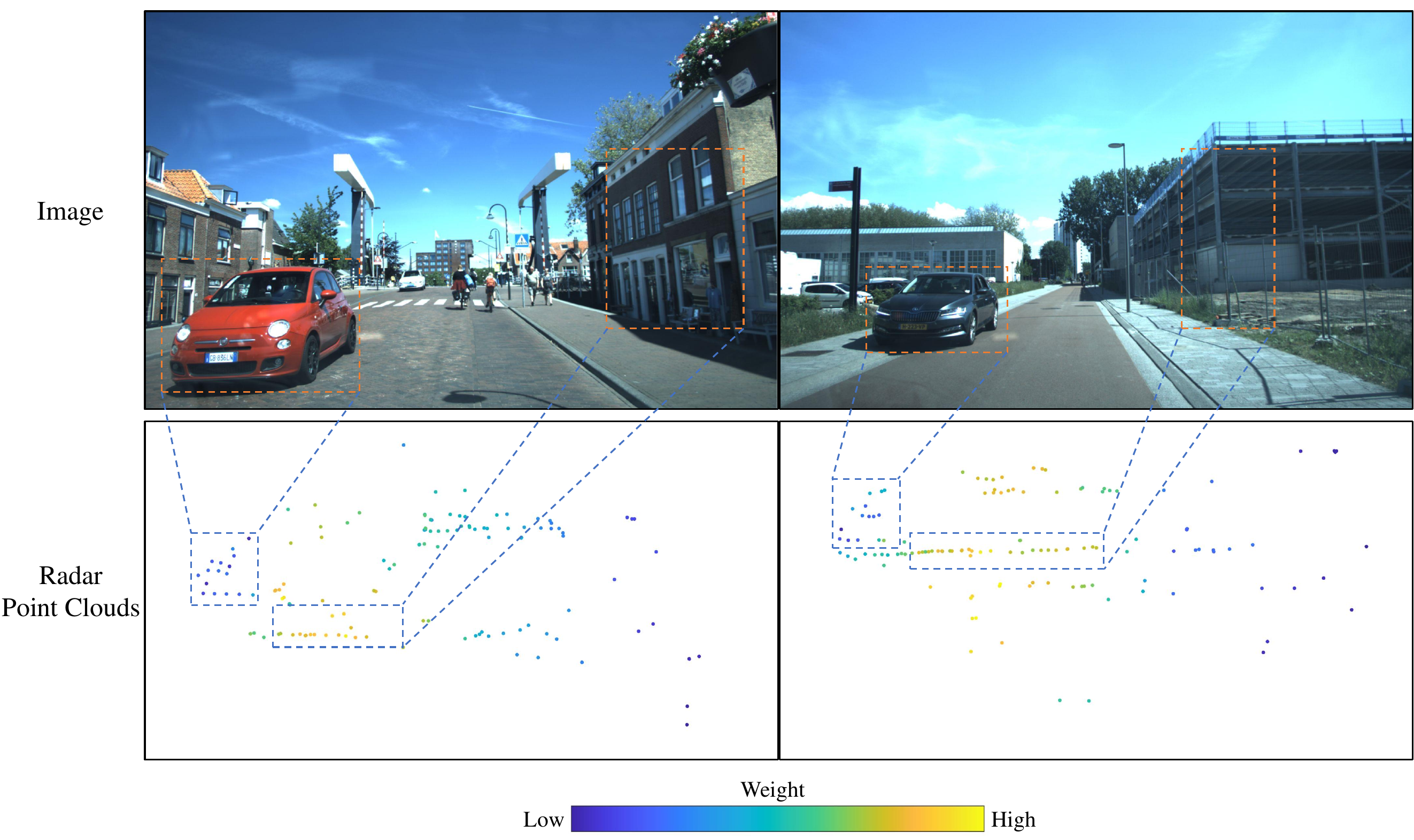}
	\caption{Visualization of point confidence. We visualize the 4D radar point cloud along with the corresponding image, using distinct colors to indicate the confidence level of each point. In the aforementioned examples, buildings and wire fences exhibit higher weightings, while moving cars exhibit lower weightings. Furthermore, the confidence level of each point gradually decreases as the distance increases, as greater trust is placed in closer stable points.}
	\label{fig11}
\end{figure}

We visualized the point confidence estimation results in the final pose-warp refinement layer to demonstrate the contribution of each point to the pose estimation. As shown in Fig. \ref{fig11}, points sampled from stationary objects (e.g., buildings and wire fences) exhibit a higher level of confidence compared with points sampled from moving objects (e.g., cars). Additionally, the weighting of the points gradually decreased as the distance increased because more distant points were generally unstable and unreliable, whereas closer points were considered more reliable. Thus, point confidence estimation can effectively reduce the effects of dynamic objects and outliers on the pose transformation.
\section{Conclusion}
\label{section6}
To the best of our knowledge, this study is the first to use a deep learning approach for 4D radar--visual odometry. The method was built on the PWC network architecture and implemented 4D radar--visual odometry in a hierarchical refinement from coarse to fine. In our approach, a multi-scale feature extraction network called Radar-PointNet++ was proposed for 4D radar point clouds to achieve fine-grained learning of sparse point clouds. Adaptive 4D radar--camera fusion methods were proposed to achieve adaptive alignment and full interaction of 4D radar point features and image features. A velocity-guided point-confidence estimation method was proposed to address the adverse effects of dynamic objects and outliers in the environment on odometry. Experiments on VoD and in-house datasets demonstrated the effectiveness of the proposed method. It is noteworthy that the proposed method achieves the localization accuracy of a 64-line LiDAR odometry. In future work, we will explore the use of unsupervised learning and incorporate an inertial measurement unit (IMU) to achieve more accurate localization results using our method.
\bibliographystyle{IEEEtran}
\bibliography{ref.bib}

\begin{thebibliography}{10}
\providecommand{\url}[1]{#1}
\csname url@samestyle\endcsname
\providecommand{\newblock}{\relax}
\providecommand{\bibinfo}[2]{#2}
\providecommand{\BIBentrySTDinterwordspacing}{\spaceskip=0pt\relax}
\providecommand{\BIBentryALTinterwordstretchfactor}{4}
\providecommand{\BIBentryALTinterwordspacing}{\spaceskip=\fontdimen2\font plus
\BIBentryALTinterwordstretchfactor\fontdimen3\font minus
  \fontdimen4\font\relax}
\providecommand{\BIBforeignlanguage}[2]{{%
\expandafter\ifx\csname l@#1\endcsname\relax
\typeout{** WARNING: IEEEtran.bst: No hyphenation pattern has been}%
\typeout{** loaded for the language `#1'. Using the pattern for}%
\typeout{** the default language instead.}%
\else
\language=\csname l@#1\endcsname
\fi
#2}}
\providecommand{\BIBdecl}{\relax}
\BIBdecl

\bibitem{orbslam3}
C.~Campos, R.~Elvira, J.~J.~G. Rodr{\'\i}guez, J.~M. Montiel, and J.~D.
  Tard{\'o}s, ``Orb-slam3: An accurate open-source library for visual,
  visual--inertial, and multimap slam,'' \emph{IEEE Transactions on Robotics},
  vol.~37, no.~6, pp. 1874--1890, 2021.

\bibitem{dmvio}
L.~Von~Stumberg and D.~Cremers, ``Dm-vio: Delayed marginalization
  visual-inertial odometry,'' \emph{IEEE Robotics and Automation Letters},
  vol.~7, no.~2, pp. 1408--1415, 2022.

\bibitem{lvisam}
T.~Shan, B.~Englot, C.~Ratti, and D.~Rus, ``Lvi-sam: Tightly-coupled
  lidar-visual-inertial odometry via smoothing and mapping,'' in \emph{2021
  IEEE international conference on robotics and automation (ICRA)}.\hskip 1em
  plus 0.5em minus 0.4em\relax IEEE, 2021, pp. 5692--5698.

\bibitem{fastlio2}
W.~Xu, Y.~Cai, D.~He, J.~Lin, and F.~Zhang, ``Fast-lio2: Fast direct
  lidar-inertial odometry,'' \emph{IEEE Transactions on Robotics}, vol.~38,
  no.~4, pp. 2053--2073, 2022.

\bibitem{sdvloam}
Z.~Yuan, Q.~Wang, K.~Cheng, T.~Hao, and X.~Yang, ``Sdv-loam: Semi-direct
  visual-lidar odometry and mapping,'' \emph{IEEE Transactions on Pattern
  Analysis and Machine Intelligence}, 2023.

\bibitem{ccvo}
T.~Zhang, N.~Li, G.~Gong, C.~Yang, G.~Hou, and X.~Lin, ``Ccvo: Cascaded cnns
  for fast monocular visual odometry towards the dynamic environment,''
  \emph{IEEE Robotics and Automation Letters}, vol.~8, no.~5, pp. 2938--2945,
  2022.

\bibitem{deepvo}
S.~Wang, R.~Clark, H.~Wen, and N.~Trigoni, ``Deepvo: Towards end-to-end visual
  odometry with deep recurrent convolutional neural networks,'' in \emph{2017
  IEEE international conference on robotics and automation (ICRA)}.\hskip 1em
  plus 0.5em minus 0.4em\relax IEEE, 2017, pp. 2043--2050.

\bibitem{lodonet}
C.~Zheng, Y.~Lyu, M.~Li, and Z.~Zhang, ``Lodonet: A deep neural network with 2d
  keypoint matching for 3d lidar odometry estimation,'' in \emph{Proceedings of
  the 28th ACM International Conference on Multimedia}, 2020, pp. 2391--2399.

\bibitem{lonet}
Q.~Li, S.~Chen, C.~Wang, X.~Li, C.~Wen, M.~Cheng, and J.~Li, ``Lo-net: Deep
  real-time lidar odometry,'' in \emph{Proceedings of the IEEE/CVF Conference
  on Computer Vision and Pattern Recognition}, 2019, pp. 8473--8482.

\bibitem{li2021self}
B.~Li, M.~Hu, S.~Wang, L.~Wang, and X.~Gong, ``Self-supervised visual-lidar
  odometry with flip consistency,'' in \emph{Proceedings of the IEEE/CVF Winter
  Conference on Applications of Computer Vision}, 2021, pp. 3844--3852.

\bibitem{hvlo}
E.~Aydemir, N.~Fetic, and M.~Unel, ``H-vlo: hybrid lidar-camera fusion for
  self-supervised odometry,'' in \emph{2022 IEEE/RSJ International Conference
  on Intelligent Robots and Systems (IROS)}.\hskip 1em plus 0.5em minus
  0.4em\relax IEEE, 2022, pp. 3302--3307.

\bibitem{rcfusion}
L.~Zheng, S.~Li, B.~Tan, L.~Yang, S.~Chen, L.~Huang, J.~Bai, X.~Zhu, and Z.~Ma,
  ``Rcfusion: Fusing 4d radar and camera with bird’s-eye view features for 3d
  object detection,'' \emph{IEEE Transactions on Instrumentation and
  Measurement}, 2023.

\bibitem{raflow}
F.~Ding, Z.~Pan, Y.~Deng, J.~Deng, and C.~X. Lu, ``Self-supervised scene flow
  estimation with 4-d automotive radar,'' \emph{IEEE Robotics and Automation
  Letters}, vol.~7, no.~3, pp. 8233--8240, 2022.

\bibitem{cmflow}
F.~Ding, A.~Palffy, D.~M. Gavrila, and C.~X. Lu, ``Hidden gems: 4d radar scene
  flow learning using cross-modal supervision,'' in \emph{Proceedings of the
  IEEE/CVF Conference on Computer Vision and Pattern Recognition}, 2023, pp.
  9340--9349.

\bibitem{iRIOM}
Y.~Zhuang, B.~Wang, J.~Huai, and M.~Li, ``4d iriom: 4d imaging radar inertial
  odometry and mapping,'' \emph{IEEE Robotics and Automation Letters}, 2023.

\bibitem{vod}
A.~Palffy, E.~Pool, S.~Baratam, J.~F.~P. Kooij, and D.~M. Gavrila,
  ``Multi-class road user detection with 3+1d radar in the view-of-delft
  dataset,'' \emph{IEEE Robotics and Automation Letters}, vol.~7, no.~2, pp.
  4961--4968, 2022.

\bibitem{li2022overview}
S.~Li, D.~Zhang, Y.~Xian, B.~Li, T.~Zhang, and C.~Zhong, ``Overview of deep
  learning application on visual slam,'' \emph{Displays}, p. 102298, 2022.

\bibitem{konda2015learning}
K.~R. Konda and R.~Memisevic, ``Learning visual odometry with a convolutional
  network.'' \emph{VISAPP (1)}, vol. 2015, pp. 486--490, 2015.

\bibitem{posenet}
A.~Kendall, M.~Grimes, and R.~Cipolla, ``Posenet: A convolutional network for
  real-time 6-dof camera relocalization,'' in \emph{Proceedings of the IEEE
  international conference on computer vision}, 2015, pp. 2938--2946.

\bibitem{szegedy2015going}
C.~Szegedy, W.~Liu, Y.~Jia, P.~Sermanet, S.~Reed, D.~Anguelov, D.~Erhan,
  V.~Vanhoucke, and A.~Rabinovich, ``Going deeper with convolutions,'' in
  \emph{Proceedings of the IEEE conference on computer vision and pattern
  recognition}, 2015, pp. 1--9.

\bibitem{espvo}
S.~Wang, R.~Clark, H.~Wen, and N.~Trigoni, ``End-to-end, sequence-to-sequence
  probabilistic visual odometry through deep neural networks,'' \emph{The
  International Journal of Robotics Research}, vol.~37, no. 4-5, pp. 513--542,
  2018.

\bibitem{tartanvo}
W.~Wang, Y.~Hu, and S.~Scherer, ``Tartanvo: A generalizable learning-based
  vo,'' in \emph{Conference on Robot Learning}.\hskip 1em plus 0.5em minus
  0.4em\relax PMLR, 2021, pp. 1761--1772.

\bibitem{xue2019beyond}
F.~Xue, X.~Wang, S.~Li, Q.~Wang, J.~Wang, and H.~Zha, ``Beyond tracking:
  Selecting memory and refining poses for deep visual odometry,'' in
  \emph{Proceedings of the IEEE/CVF conference on computer vision and pattern
  recognition}, 2019, pp. 8575--8583.

\bibitem{liu2021unsupervised}
Y.~Liu, H.~Wang, J.~Wang, and X.~Wang, ``Unsupervised monocular visual odometry
  based on confidence evaluation,'' \emph{IEEE Transactions on Intelligent
  Transportation Systems}, vol.~23, no.~6, pp. 5387--5396, 2021.

\bibitem{teed2022deep}
Z.~Teed, L.~Lipson, and J.~Deng, ``Deep patch visual odometry,'' \emph{arXiv
  preprint arXiv:2208.04726}, 2022.

\bibitem{cross-modal}
B.~Li, S.~Wang, H.~Ye, X.~Gong, and Z.~Xiang, ``Cross-modal knowledge
  distillation for depth privileged monocular visual odometry,'' \emph{IEEE
  Robotics and Automation Letters}, vol.~7, no.~3, pp. 6171--6178, 2022.

\bibitem{nicolai2016deep}
A.~Nicolai, R.~Skeele, C.~Eriksen, and G.~A. Hollinger, ``Deep learning for
  laser based odometry estimation,'' in \emph{RSS workshop Limits and
  Potentials of Deep Learning in Robotics}, vol. 184, 2016, p.~1.

\bibitem{deeppco}
W.~Wang, M.~R.~U. Saputra, P.~Zhao, P.~Gusmao, B.~Yang, C.~Chen, A.~Markham,
  and N.~Trigoni, ``Deeppco: End-to-end point cloud odometry through deep
  parallel neural network,'' in \emph{2019 IEEE/RSJ International Conference on
  Intelligent Robots and Systems (IROS)}.\hskip 1em plus 0.5em minus
  0.4em\relax IEEE, 2019, pp. 3248--3254.

\bibitem{dmlo}
Z.~Li and N.~Wang, ``Dmlo: Deep matching lidar odometry,'' in \emph{2020
  IEEE/RSJ International Conference on Intelligent Robots and Systems
  (IROS)}.\hskip 1em plus 0.5em minus 0.4em\relax IEEE, 2020, pp. 6010--6017.

\bibitem{xu2022robust}
Y.~Xu, J.~Lin, J.~Shi, G.~Zhang, X.~Wang, and H.~Li, ``Robust self-supervised
  lidar odometry via representative structure discovery and 3d inherent error
  modeling,'' \emph{IEEE Robotics and Automation Letters}, vol.~7, no.~2, pp.
  1651--1658, 2022.

\bibitem{wang2021pwclo}
G.~Wang, X.~Wu, Z.~Liu, and H.~Wang, ``Pwclo-net: Deep lidar odometry in 3d
  point clouds using hierarchical embedding mask optimization,'' in
  \emph{Proceedings of the IEEE/CVF conference on computer vision and pattern
  recognition}, 2021, pp. 15\,910--15\,919.

\bibitem{wang2022efficient}
G.~Wang, X.~Wu, S.~Jiang, Z.~Liu, and H.~Wang, ``Efficient 3d deep lidar
  odometry,'' \emph{IEEE Transactions on Pattern Analysis and Machine
  Intelligence}, vol.~45, no.~5, pp. 5749--5765, 2022.

\bibitem{deng2023nerf}
J.~Deng, X.~Chen, S.~Xia, Z.~Sun, G.~Liu, W.~Yu, and L.~Pei, ``Nerf-loam:
  Neural implicit representation for large-scale incremental lidar odometry and
  mapping,'' \emph{arXiv preprint arXiv:2303.10709}, 2023.

\bibitem{yu2022stcloc}
S.~Yu, C.~Wang, Y.~Lin, C.~Wen, M.~Cheng, and G.~Hu, ``Stcloc: Deep lidar
  localization with spatio-temporal constraints,'' \emph{IEEE Transactions on
  Intelligent Transportation Systems}, vol.~24, no.~1, pp. 489--500, 2022.

\bibitem{zhang2014real}
J.~Zhang, M.~Kaess, and S.~Singh, ``Real-time depth enhanced monocular
  odometry,'' in \emph{2014 IEEE/RSJ International Conference on Intelligent
  Robots and Systems}.\hskip 1em plus 0.5em minus 0.4em\relax IEEE, 2014, pp.
  4973--4980.

\bibitem{zhang2015visual}
J.~Zhang and S.~Singh, ``Visual-lidar odometry and mapping: Low-drift, robust,
  and fast,'' in \emph{2015 IEEE International Conference on Robotics and
  Automation (ICRA)}.\hskip 1em plus 0.5em minus 0.4em\relax IEEE, 2015, pp.
  2174--2181.

\bibitem{huang2020lidar}
S.-S. Huang, Z.-Y. Ma, T.-J. Mu, H.~Fu, and S.-M. Hu, ``Lidar-monocular visual
  odometry using point and line features,'' in \emph{2020 IEEE International
  Conference on Robotics and Automation (ICRA)}.\hskip 1em plus 0.5em minus
  0.4em\relax IEEE, 2020, pp. 1091--1097.

\bibitem{limo}
J.~Graeter, A.~Wilczynski, and M.~Lauer, ``Limo: Lidar-monocular visual
  odometry,'' in \emph{2018 IEEE/RSJ international conference on intelligent
  robots and systems (IROS)}.\hskip 1em plus 0.5em minus 0.4em\relax IEEE,
  2018, pp. 7872--7879.

\bibitem{sdv}
Z.~Yuan, Q.~Wang, K.~Cheng, T.~Hao, and X.~Yang, ``Sdv-loam: Semi-direct
  visual-lidar odometry and mapping,'' \emph{IEEE Transactions on Pattern
  Analysis and Machine Intelligence}, 2023.

\bibitem{yue2020lidar}
J.~Yue, W.~Wen, J.~Han, and L.-T. Hsu, ``Lidar data enrichment using deep
  learning based on high-resolution image: An approach to achieve
  high-performance lidar slam using low-cost lidar,'' \emph{arXiv preprint
  arXiv:2008.03694}, 2020.

\bibitem{an2022visual}
Y.~An, J.~Shi, D.~Gu, and Q.~Liu, ``Visual-lidar slam based on unsupervised
  multi-channel deep neural networks,'' \emph{Cognitive Computation}, vol.~14,
  no.~4, pp. 1496--1508, 2022.

\bibitem{he2016deep}
K.~He, X.~Zhang, S.~Ren, and J.~Sun, ``Deep residual learning for image
  recognition,'' in \emph{Proceedings of the IEEE conference on computer vision
  and pattern recognition}, 2016, pp. 770--778.

\bibitem{pointnet++}
C.~R. Qi, L.~Yi, H.~Su, and L.~J. Guibas, ``Pointnet++: Deep hierarchical
  feature learning on point sets in a metric space,'' \emph{Advances in neural
  information processing systems}, vol.~30, 2017.

\bibitem{pointpwc}
W.~Wu, Z.~Y. Wang, Z.~Li, W.~Liu, and L.~Fuxin, ``Pointpwc-net: Cost volume on
  point clouds for (self-) supervised scene flow estimation,'' in
  \emph{Computer Vision--ECCV 2020: 16th European Conference, Glasgow, UK,
  August 23--28, 2020, Proceedings, Part V 16}.\hskip 1em plus 0.5em minus
  0.4em\relax Springer, 2020, pp. 88--107.

\bibitem{besl1992method}
P.~J. Besl and N.~D. McKay, ``Method for registration of 3-d shapes,'' in
  \emph{Sensor fusion IV: control paradigms and data structures}, vol.
  1611.\hskip 1em plus 0.5em minus 0.4em\relax Spie, 1992, pp. 586--606.

\bibitem{chen1992object}
Y.~Chen and G.~Medioni, ``Object modelling by registration of multiple range
  images,'' \emph{Image and vision computing}, vol.~10, no.~3, pp. 145--155,
  1992.

\bibitem{generalized}
A.~Segal, D.~Haehnel, and S.~Thrun, ``Generalized-icp.'' in \emph{Robotics:
  science and systems}, vol.~2, no.~4.\hskip 1em plus 0.5em minus 0.4em\relax
  Seattle, WA, 2009, p. 435.

\bibitem{stoyanov2012fast}
T.~Stoyanov, M.~Magnusson, H.~Andreasson, and A.~J. Lilienthal, ``Fast and
  accurate scan registration through minimization of the distance between
  compact 3d ndt representations,'' \emph{The International Journal of Robotics
  Research}, vol.~31, no.~12, pp. 1377--1393, 2012.

\bibitem{zhang2017low}
J.~Zhang and S.~Singh, ``Low-drift and real-time lidar odometry and mapping,''
  \emph{Autonomous Robots}, vol.~41, pp. 401--416, 2017.

\end{thebibliography}
\end{document}